# On Reasonable and Forced Goal Orderings and their Use in an Agenda-Driven Planning Algorithm

**Jana Koehler**                                                                 JANA_KOEHLER@CH.SCHINDLER.COM
*Schindler Lifts, Ltd.*
*R & D Technology Management*
*6031 Ebikon, Switzerland*

**Jörg Hoffmann**                                                                HOFFMANN@INFORMATIK.UNI-FREIBURG.DE
*Institute for Computer Science*
*Albert Ludwigs University*
*Georges-Köhler-Allee, Geb. 52*
*79110 Freiburg, Germany*

## Abstract

The paper addresses the problem of computing goal orderings, which is one of the longstanding issues in AI planning. It makes two new contributions. First, it formally defines and discusses two different goal orderings, which are called the *reasonable* and the *forced* ordering. Both orderings are defined for simple STRIPS operators as well as for more complex ADL operators supporting negation and conditional effects. The complexity of these orderings is investigated and their practical relevance is discussed. Secondly, two different methods to compute reasonable goal orderings are developed. One of them is based on planning graphs, while the other investigates the set of actions directly. Finally, it is shown how the ordering relations, which have been derived for a given set of goals $\mathcal{G}$, can be used to compute a so-called *goal agenda* that divides $\mathcal{G}$ into an ordered set of subgoals. Any planner can then, in principle, use the goal agenda to plan for increasing sets of subgoals. This can lead to an exponential complexity reduction, as the solution to a complex planning problem is found by solving easier subproblems. Since only a polynomial overhead is caused by the goal agenda computation, a potential exists to dramatically speed up planning algorithms as we demonstrate in the empirical evaluation, where we use this method in the IPP planner.

## 1. Introduction

How to effectively plan for interdependent subgoals has been in the focus of AI planning research for a very long time. Starting with the early work on ABSTRIPS (Sacerdoti, 1974) or on conjunctive-goal planning problems (Chapman, 1987), quite a number of approaches have been presented and the complexity of the problems has been studied. But until today, planners have made only some progress in solving bigger planning instances and scalability of classical planning systems is still a problem.

In this paper, we focus on the following problem: Given a set of conjunctive goals, can we define and detect an ordering relation over subsets from the original goal set? To arrive at such an ordering relation over subsets, we first focus on the atomic facts contained in the goal set. We formally define two closely related ordering relations over such atomic goals, which we call *reasonable* and *forced* ordering, and study their complexity. It turns out that both are very hard to decide.

Consequently, we introduce two efficient methods that can both be used to approximate *reasonable* goal orderings. The definitions are first given for simple STRIPS domains, where the desired theoretical properties can be easily proven. Afterwards, we extend our definitions to ADL operators (Pednault, 1989) handling conditional effects and negative preconditions, and discuss why we do not further invest any effort in trying to find *forced* orderings.

We show how a set of ordering relations between atomic goals can be used to divide the goal set into disjunct subsets, and how these subsets can be ordered with respect to each other. The resulting sequence of subsets comprises the so-called *goal agenda*, which can then be used to control an agenda-driven planning algorithm.

The method, called **G**oal **A**genda **M**anager, is implemented in the context of the IPP planning system, where we show its potential of exponentially reducing computation times on certain planning domains.

The paper is organized as follows: Section 2 introduces and motivates reasonable and forced goal orderings. Starting with simple STRIPS operators, they are formally defined, and their complexity is investigated. In Section 3, we present two methods, which compute an approximation of the reasonable ordering and discuss both orderings from a more practical point of view. The section concludes with an extension of our definitions to ADL operators having conditional effects. Section 4 shows how a planning system can benefit from ordering information by computing a goal agenda that guides the planner. We define how subsets of goals can be ordered with respect to each other and discuss how a goal agenda can affect the theoretical properties, in particular the completeness of a planning algorithm. Section 5 contains the empirical evaluation of our work, showing results that we obtained using the goal agenda in IPP. In Section 6 we summarize our approach in the light of related work. The paper concludes with an outlook on possible future research directions in Section 7.

## 2. Ordering Relations between Atomic Goals

For a start, we only investigate simple STRIPS domains just allowing sets of atoms to describe states, the preconditions, and the add and delete lists of operators.

**Definition 1 (State)** *The set of all ground atoms is denoted with $P$. A state $s \in 2^P$ is a subset of ground atoms.*

Note that all states are assumed to be *complete*, i.e., we always know for an atom $p$ whether $p \in s$ or $p \notin s$ holds. We also assume that all operator schemata are ground, i.e., we only talk about *actions*.

**Definition 2 (Strips Action)** *A STRIPS action $o$ has the usual form*

$$pre(o) \longrightarrow ADD \ add(o) \ DEL \ del(o)$$

*where $pre(o)$ are the preconditions of $o$, $add(o)$ is the Add list of $o$ and $del(o)$ is the Delete list of the action, each being a set of ground atoms. We also assume that $del(o) \cap add(o) = \emptyset$. The result of applying a STRIPS action to a state is defined as usual:*

$$Result(s, o) := \begin{cases} (s \cup add(o)) \setminus del(o) & \text{if } pre(o) \subseteq s \\ s & \text{otherwise} \end{cases}$$

If $pre(o) \subseteq s$ holds, the action is said to be *applicable in $s$*. The result of applying a sequence of more than one action to a state is recursively defined as

$$Result(s, \langle o_1, \ldots, o_n \rangle) := Result(Result(s, \langle o_1, \ldots, o_{n-1} \rangle), o_n).$$

**Definition 3 (Planning Problem)** *A planning problem $(\mathcal{O}, \mathcal{I}, \mathcal{G})$ is a triple where $\mathcal{O}$ is the set of actions, and $\mathcal{I}$ (the initial state) and $\mathcal{G}$ (the goals) are sets of ground atoms. A plan $\mathcal{P}$ is an ordered sequence of actions. If all actions in a plan are taken out of a certain action set $\mathcal{O}$, we denote this by writing $\mathcal{P}^{\mathcal{O}}$.*

Note that we define a plan to be a sequence of *actions*, not a sequence of *parallel steps*, as it is done for GRAPHPLAN (Blum & Furst, 1997), for example. This makes the subsequent theoretical investigation more readable. The results directly carry over to parallel plans.

Given two atomic goals $A$ and $B$, various ways to define an ordering relation over them can be imagined. First, one can distinguish between *domain-specific* and *domain-independent* goal ordering relations. But although domain-specific orderings can be very effective, they need to be redeveloped for each single domain. Therefore, one is in particular interested in domain-independent ordering relations having a broader range of applicability. Secondly, following Hüllem et al. (1999), one can distinguish the *goal selection* and the *goal achievement* order. The first ordering determines in which order a planner works on the various atomic goals, while the second one determines the order, in which the solution plan achieves the goals. In this paper, we compute an ordering of the latter type. In the agenda-driven planning approach that we propose later in the paper, both orderings coincide anyway. The goals that are achieved first in the plan are those that the planner works on first.

The following scenario motivates how an achievement order for goals can be possibly defined. Given two atomic goals $A$ and $B$, for which a solution plan exists, let us assume the planner has just achieved the goal $A$, i.e., it has arrived at a state $s_{(A, \neg B)}$, in which $A$ holds, but $B$ does not hold yet. Now, if there exists a plan that is executable in $s_{(A, \neg B)}$ and achieves $B$ without ever deleting $A$, a solution has been found. If no such plan can be found, then two possible reasons exist:

1. The problem is unsolvable—achieving $A$ first leads the planner into a deadlock situation. Thus, the planner is *forced* to achieve $B$ before or simultaneously with $A$.

2. The only existing solution plans have to destroy $A$ temporarily in order to achieve $B$. But then, $A$ should not be achieved first. Instead, it seems to be *reasonable* to achieve $B$ before or simultaneously with $A$ for the sake of shorter solution plans.

In the first situation, the ordering "$B$ before or simultaneously with $A$" is *forced* by inherent properties of the planning domain. In the second situation, the ordering "$B$ before or simultaneously with $A$" appears to be *reasonable* in order to avoid non-optimal plans. Consequently, we will define two goal orderings, called the *forced* and the *reasonable* ordering. For the sake of clarity, we first give some more basic definitions.

KOEHLER & HOFFMANN**Definition 4 (Reachable State)** *Let $(\mathcal{O}, \mathcal{I}, \mathcal{G})$ be a planning problem and let $P$ be the set of ground atoms that occur in the problem. We say that a state $s \subseteq P$ is* reachable, *iff there exists a sequence $\langle o_1, \ldots, o_n \rangle$ out of actions in $\mathcal{O}$ for which $s = Result(\mathcal{I}, \langle o_1, \ldots, o_n \rangle)$ holds.*

**Definition 5 (Generic State $s_{(A, \neg B)}$)** *Let $(\mathcal{O}, \mathcal{I}, \mathcal{G})$ be a planning problem. By $s_{(A, \neg B)}$ we denote any reachable state in which $A$ has just been achieved, but $B$ is* FALSE, *i.e., $B \notin s_{(A, \neg B)}$ and there is a sequence of actions $\langle o_1, \ldots, o_n \rangle$ such that $s_{(A, \neg B)} = Result(\mathcal{I}, \langle o_1, \ldots, o_n \rangle)$, with $A \in add(o_n)$.*

One can imagine $s_{(A, \neg B)}$ as a state about which we only have incomplete information. All the states $s$ it represents satisfy $s \models A, \neg B$, but the other atoms $p \in P$ with $p \neq A, B$ can adopt arbitrary truth values.

**Definition 6 (Reduced Action Set $\mathcal{O}_A$)** *Let $(\mathcal{O}, \mathcal{I}, \mathcal{G})$ be a planning problem, and let $A \in \mathcal{G}$ be an atomic goal. By $\mathcal{O}_A$ we denote the set of all actions that do not delete $A$, i.e., $\mathcal{O}_A = \{o \in \mathcal{O} \mid A \notin del(o)\}$.*

We are now prepared to define what we exactly mean by *forced* and *reasonable* goal orderings.

**Definition 7 (Forced Ordering $\leq_f$)** *Let $(\mathcal{O}, \mathcal{I}, \mathcal{G})$ be a planning problem, and let $A, B \in \mathcal{G}$ be two atomic goals. We say that there is a* forced ordering *between $B$ and $A$, written $B \leq_f A$, if and only if*

$$\forall s_{(A, \neg B)}: \quad \neg \exists \, \mathcal{P}^\mathcal{O}: \, B \in Result(s_{(A, \neg B)}, \mathcal{P}^\mathcal{O})$$

If Definition 7 is satisfied, then each plan achieving $A$ and $B$ must achieve $B$ before or simultaneously with $A$, because otherwise it will encounter a deadlock, rendering the problem unsolvable.

**Definition 8 (Reasonable Ordering $\leq_r$)** *Let $(\mathcal{O}, \mathcal{I}, \mathcal{G})$ be a planning problem, and let $A, B \in \mathcal{G}$ be two atomic goals. We say that there is a* reasonable ordering *between $B$ and $A$, written $B \leq_r A$, if and only if*

$$\forall s_{(A, \neg B)}: \quad \neg \exists \, \mathcal{P}^{\mathcal{O}_A}: \, B \in Result(s_{(A, \neg B)}, \mathcal{P}^{\mathcal{O}_A})$$

Definition 8 gives $B \leq_r A$ the meaning that if, after the goal $A$ has been achieved, there is no plan anymore that achieves $B$ without—at least temporarily—destroying $A$, then $B$ is a goal prior to $A$.

We remark that obviously $B \leq_f A$ implies $B \leq_r A$, but not vice versa. We also make a slightly less obvious observation at this point: The formulae in Definitions 7 and 8 use a universal quantification over states $s_{(A, \neg B)}$. If in a planning problem there is no such state at all, the formulae are satisfied and the goals $A$ and $B$ get ordered, i.e., $B \leq_f A$ and $B \leq_r A$ follow, respectively. In this case, however, there is not much information gained by a goal ordering between $A$ and $B$, because *any* sequence of actions will achieve $B$ prior or simultaneously with $A$—$A$ cannot be achieved with $B$ still being FALSE. Thus in this case, the ordering relations $B \leq_f A$ and $B \leq_r A$ are *trivial* in the sense that no reasonable planner would invest much effort in considering the goals $A$ and $B$ ordered the other way round anyway.

342

**Definition 9 (Trivial Ordering Relation)** *Let $(\mathcal{O}, \mathcal{I}, \mathcal{G})$ be a planning problem, and let $A, B \in \mathcal{G}$ be two atomic goals. An ordering relation $B \leq_f A$ or $B \leq_r A$ is called* trivial *iff there is no state $s_{(A, \neg B)}$.*

In this paper, we will usually consider forced and reasonable goal orderings as non-trivial orderings and make the distinction explicit only if we have to do so.

Definitions 7 and 8 seem to deliver promising candidates for an achievement order. Unfortunately, both are very hard to test: it turns out that their corresponding decision problems are PSPACE hard.

**Theorem 1** *Let* F_ORDER *denote the following problem:*

*Given two atomic facts $A$ and $B$, as well as an action set $\mathcal{O}$ and an initial state $\mathcal{I}$, does $B \leq_f A$ hold ?*

*Deciding* F_ORDER *is* PSPACE-*hard.*

**Proof:** The proof proceeds by polynomially reducing PLANSAT (Bylander, 1994)—the decision problem of whether there exists a solution plan for a given arbitrary STRIPS planning instance—to the problem of deciding F_ORDER.

Let $\mathcal{I}$, $\mathcal{G}$, and $\mathcal{O}$ denote the initial state, the goal state, and the action set in an arbitrary STRIPS instance. Let $A$, $B$, and $C$ be new atomic facts not contained in the instance so far. We build a new action set and initial state for our F_ORDER instance by setting

$$\mathcal{O}' := \mathcal{O} \cup \left\{ \begin{array}{lll} o_{I_1} & = \{C\} & \longrightarrow \text{ADD } \{A\} \text{ DEL } \{C\}, \\ o_{I_2} & = \{A\} & \longrightarrow \text{ADD } \mathcal{I} \text{ DEL } \{A\}, \\ o_G & = \mathcal{G} & \longrightarrow \text{ADD } \{B\} \text{ DEL } \emptyset \end{array} \right\}$$

and

$$\mathcal{I}' := \{C\}$$

With these definitions, reaching $B$ from $A$ is equivalent to solving the original problem. The other way round, unreachability of $B$ from $A$—forced ordering $B \leq_f A$—is equivalent to the unsolvability of the original problem. In order to prove this, we consider the following: The only way of achieving $A$ is by applying $o_{I_1}$ to $\mathcal{I}'$. Consequently, the only state $s_{(A, \neg B)}$ is $\{A\}$, cf. Definition 5. Thus starting with the assumption that $B \leq_f A$ is valid, we apply the following equivalences:

$\quad B \leq_f A$
$\Leftrightarrow \quad \forall s_{(A, \neg B)} : \neg \exists \, \mathcal{P}^{\mathcal{O}'} : B \in Result(s_{(A, \neg B)}, \mathcal{P}^{\mathcal{O}'})$ \hfill cf. Definition 7
$\Leftrightarrow \quad \neg \exists \, \mathcal{P}^{\mathcal{O}'} : B \in Result(\{A\}, \mathcal{P}^{\mathcal{O}'})$ \hfill $\{A\}$ is the only reachable state $s_{(A, \neg B)}$
$\Leftrightarrow \quad \neg \exists \, \mathcal{P}^{\mathcal{O}} : \mathcal{G} \subseteq Result(\mathcal{I}, \mathcal{P}^{\mathcal{O}})$ \hfill with the definition of $\mathcal{O}'$
$\Leftrightarrow \quad$ no solution plan exists for $\mathcal{I}, \mathcal{G}$ and given $\mathcal{O}$

Thus, the complement of PLANSAT can be polynomially reduced to F_ORDER. As PSPACE = co-PSPACE, we are done. ∎

**Theorem 2** *Let R_ORDER denote the following problem:*

Given two atomic facts $A$ and $B$, as well as an action set $\mathcal{O}$ and an initial state $\mathcal{I}$, does $B \leq_r A$ hold?

*Deciding R_ORDER is PSPACE-hard.*

**Proof:** The proof proceeds by polynomially reducing PLANSAT to R_ORDER.

Let $\mathcal{I}$, $\mathcal{G}$, and $\mathcal{O}$ be the initial state, the goal state, and the action set in an arbitrary STRIPS planning instance. Let $A$, $B$, $C$, and $D$ be new atomic facts not contained in the instance so far. We define the new action set $\mathcal{O}'$ by setting

$$\mathcal{O}' := \mathcal{O} \cup \left\{ \begin{array}{lll} o_{I_1} &= \{C\} &\longrightarrow \text{ADD } \{A,D\} \text{ DEL } \{C\}, \\ o_{I_2} &= \{A,D\} &\longrightarrow \text{ADD } \mathcal{I} \text{ DEL } \{D\}, \\ o_G &= \mathcal{G} &\longrightarrow \text{ADD } \{B\} \text{ DEL } \emptyset \end{array} \right\}$$

and the new initial state by

$$\mathcal{I}' := \{C\}$$

As in the proof of Theorem 1, the intention behind these definitions is to make solvability of the original problem equivalent to reachability of $B$ from $A$. For reasonable orderings, reachability is concerned with actions that do not delete $A$, which is why we need the safety condition $D$.

Precisely, the only way to achieve $A$ is by applying $o_{I_1}$ to $\mathcal{I}'$, i.e., per Definition 5 the only state $s_{(A,\neg B)}$ is $\{A,D\}$. As no action in the new operator set $\mathcal{O}'$ deletes $A$, we have the following sequence of equivalences.

$$\begin{array}{rll} & B \leq_r A & \\ \Leftrightarrow & \forall s_{(A,\neg B)} \; \neg \exists \, \mathcal{P}^{\mathcal{O}'_A} : B \in Result(s_{(A,\neg B)}, \mathcal{P}^{\mathcal{O}'_A}) & \text{cf. Definition 8} \\ \Leftrightarrow & \neg \exists \, \mathcal{P}^{\mathcal{O}'_A} : \; B \in Result(\{A,D\}, \mathcal{P}^{\mathcal{O}'_A}) & \{A,D\} \text{ is the only reachable state } s_{(A,\neg B)} \\ \Leftrightarrow & \neg \exists \, \mathcal{P}^{\mathcal{O}'} : \; B \in Result(\{A,D\}, \mathcal{P}^{\mathcal{O}'}) & \text{no action in } \mathcal{O}' \text{ deletes } A \\ \Leftrightarrow & \neg \exists \, \mathcal{P}^{\mathcal{O}} \text{ such that } \; \mathcal{G} \subseteq Result(\mathcal{I}, \mathcal{P}^{\mathcal{O}}) & \text{with the definition of } \mathcal{O}' \\ \Leftrightarrow & \text{no solution plan exists for } \mathcal{I}, \mathcal{G}, \mathcal{O} & \end{array}$$

Thus, the complement of PLANSAT can be polynomially reduced to R_ORDER. With PSPACE = co-PSPACE, we are done. ∎

Consequently, finding reasonable and forced ordering relations between atomic goals is already as hard as the original planning problem and it appears unlikely that a planner will gain any advantage from doing that. A possible way out of the dilemma is to define new ordering relations, which can be decided in polynomial time and which are, ideally, sufficient for the existence of reasonable or forced goal orderings. In the following, we introduce two such orderings.

## 3. The Computation of Goal Orderings

In this section, we will

1. define a goal ordering $\leq_e$, which can be computed using GRAPHPLAN's exclusivity information about facts. We prove that this ordering is sufficient for $\leq_r$ and that it can be decided in polynomial time (the subscript "e" stands for "efficient").

2. define a goal ordering $\leq_h$, which is computed based on a heuristic method that is much faster than the computation based on GRAPHPLAN, and also delivers powerful goal ordering information (the subscript "h" stands for "heuristic").

3. discuss that most of the currently available benchmark planning domains do not contain forced orderings, i.e., $\leq_f$ will fail in providing a problem decomposition for them.

4. show how our orderings can be extended to handle more expressive ADL operators.

### 3.1 Reasonable Goal Orderings based on GRAPHPLAN

A goal ordering is always computed for a specific planning problem involving an initial state $\mathcal{I}$, a goal set $\mathcal{G} \supseteq \{A, B\}$, and the set $\mathcal{O}$ of all ground actions. In order to develop an efficient computational method, we proceed in two steps now:

1. We compute more knowledge about the generic state $s_{(A,\neg B)}$.

2. We define the relation $\leq_e$ and investigate its theoretical properties. In particular, we prove that $\leq_e$ implies $\leq_r$.

The state $s_{(A,\neg B)}$ represents states that are reachable from $\mathcal{I}$, and in which $A$ has been achieved, but $B$ does not hold. Given this information about $s_{(A,\neg B)}$, one can derive additional knowledge about it. In particular, it is possible to determine a subset of atoms F, of which one definitely knows that $\mathsf{F} \cap s_{(A,\neg B)} = \emptyset$ must hold. One method to determine F is obtained via the computation of invariants, i.e., logical formulae that hold in all reachable states, cf. (Fox & Long, 1998). After having determined the invariants, one assumes that $A$ holds, but $B$ does not, and then computes the logical implications. Another possibility is to simply use GRAPHPLAN (Blum & Furst, 1997). Starting from $\mathcal{I}$ with $\mathcal{O}$, the planning graph is built until the graph has *leveled* off at some time step. The proposition level at this time step represents a set of states, which is a superset of all states that are reachable from $\mathcal{I}$ when applying actions from $\mathcal{O}$. All atoms, which are marked as *mutually exclusive* (Blum & Furst, 1997) of $A$ in this level can never hold in a state satisfying $A$. Thus, they cannot hold in $s_{(A,\neg B)}$. We denote this set with $\mathsf{F}_{GP}^A$—the False set with respect to $A$ returned by GRAPHPLAN.[1]

$$\mathsf{F}_{GP}^A := \{p \mid p \text{ is exclusive of } A \text{ when the graph has leveled off}\} \qquad (1)$$

Note that the planning graph is only grown once for a given $\mathcal{I}$ and $\mathcal{O}$, but can be used to determine the $\mathsf{F}_{GP}^A$ sets for all atomic goals $A \in \mathcal{G}$.

---

1. We assume the reader to be familiar with GRAPHPLAN, because this planning system is very well known in the planning research community. Otherwise, (Blum & Furst, 1997) provide the necessary background.

**Lemma 1** $\mathsf{F}_{GP}^A \cap s_A = \emptyset$ *holds for all states $s_A$ satisfying $A \in s_A$ that are reachable from $\mathcal{I}$ using actions from $\mathcal{O}$.*

The proof follows immediately from the definitions of "level-off" and "two propositions being mutual exclusive" given in (Blum & Furst, 1997).

We now provide a simple test which is sufficient for the existence of a reasonable ordering $B \leq_r A$ between two atomic goals $A$ and $B$.

**Definition 10 (Efficient Ordering $\leq_e$)** *Let $(\mathcal{O}, \mathcal{I}, \mathcal{G} \supseteq \{A, B\})$ be a planning problem. Let $\mathsf{F}_{GP}^A$ be the False set for $A$. The ordering $B \leq_e A$ holds if and only if*

$$\forall\, o \in \mathcal{O}_A : B \in add(o) \Rightarrow pre(o) \cap \mathsf{F}_{GP}^A \neq \emptyset$$

This means, $B$ is ordered before $A$ if the reduced action set only contains actions, which either do not have $B$ in their add lists or if they do, then they require a precondition which is contained in the False set. Such preconditions can never hold in a state satisfying $A$ and thus, these actions will never be applicable.

**Theorem 3**

$$B \leq_e A \Rightarrow B \leq_r A$$

**Proof:** Assume that $B \not\leq_r A$, i.e., $B \in Result(s_{(A, \neg B)}, \mathcal{P}^{\mathcal{O}_A})$ for a reachable state $s_{(A, \neg B)}$ with $A \in s_{(A, \neg B)}$, $B \notin s_{(A, \neg B)}$, and a Plan $\mathcal{P}^{\mathcal{O}_A} = \langle o_1, \ldots, o_n \rangle$ where $o_i \in \mathcal{O}_A$ for $1 \leq i \leq n$.

As $A \notin del(o_i)$ for all $i$ (Definition 6), we have

$$A \in Result(s_{(A, \neg B)}, \langle o_1, \ldots, o_i \rangle) \text{ for } 0 \leq i \leq n$$

and, with Lemma 1,

$$\mathsf{F}_{GP}^A \cap Result(s_{(A, \neg B)}, \langle o_1, \ldots, o_i \rangle) = \emptyset \text{ for } 0 \leq i \leq n \tag{2}$$

Furthermore, as $B \notin s_{(A, \neg B)}$, but $B \in Result(s_{(A, \neg B)}, \langle o_1, \ldots, o_n \rangle)$, there must be a step which makes $B$ true, i.e.,

$$\exists 1 \leq k \leq n : B \notin Result(s_{(A, \neg B)}, \langle o_1, \ldots, o_{k-1} \rangle) \land B \in Result(s_{(A, \neg B)}, \langle o_1, \ldots, o_k \rangle)$$

For this step, we obviously have $B \in add(o_k)$ and consequently, with the definition of $B \leq_e A$, $pre(o_k) \cap \mathsf{F}_{GP}^A \neq \emptyset$. Now, as $o_k$ must be applicable in the state where it is executed (otherwise it would not add anything to this state), the preconditions of $o_k$ must hold, i.e., $pre(o_k) \subseteq Result(s_{(A, \neg B)}, \langle o_1, \ldots, o_{k-1} \rangle)$. This immediately leads to $\mathsf{F}_{GP}^A \cap Result(s_{(A, \neg B)}, \langle o_1, \ldots, o_{k-1} \rangle) \neq \emptyset$, which is a contradiction to Equation (2). ■

Quite obviously, the ordering $\leq_e$ can be decided in polynomial time.

**Theorem 4** *Let* E_ORDER *denote the following problem:*

*Given two atomic facts $A$ and $B$, as well as an initial state $\mathcal{I}$ and an action set $\mathcal{O}$, does $B \leq_e A$ hold ?*

*Then,* E_ORDER *can be decided in polynomial time:* E_ORDER $\in$ P.

**Proof:** To begin with, we need to show that computing $\mathsf{F}_{GP}^A$ takes only polynomial time. From the results in (Blum & Furst, 1997), it follows directly that building a planning graph is polynomial in $|\mathcal{I}|$, $|\mathcal{O}|$, $l$ and $t$, where $l$ is the maximal length of any precondition, add or delete list of an action, and $t$ is the number of time steps built. Taking $l$ as a parameter of the input size, it remains to show that a planning graph levels off after a polynomial number $t$ of time steps. Now, a planning graph has leveled off if between some time steps $t$ and $t+1$ neither the set of facts nor the number of exclusion relations change. Between two subsequent time steps, the set of facts can only increase—facts already occuring in the graph remain there—and the number of exclusions can only decrease—non-exclusive facts will be non-exclusive in all subsequent layers. Thus, the maximal number of time steps to be built until the graph has leveled off is dominated by the maximal number of changes that can occur between two subsequent layers, which is dominated by the maximal number of facts plus the maximal number of exclusion relations. The maximal number of facts is $O(|\mathcal{I}| + |\mathcal{O}| * l)$, and the maximal number of exclusions is $O((|\mathcal{I}| + |\mathcal{O}| * l)^2)$, the square of the maximal number of facts.

Having computed $\mathsf{F}_{GP}^A$ in polynomial time, testing $B \leq_e A$ involves looking at all actions in $\mathcal{O}$, and rejecting them if they either

- delete $A$, which is decidable in time $O(l)$, or

- have a precondition, which is an element of $\mathsf{F}_{GP}^A$, decidable in time $O(l * (|\mathcal{I}| + |\mathcal{O}| * l))$.

Thus we have an additional runtime for the test, which is $O(|\mathcal{O}| * l * (|\mathcal{I}| + |\mathcal{O}| * l))$. ∎

Let us consider the following example, which illustrates the computation of $\leq_e$ using a common representational variant of the *blocks world* with actions to **stack**, **unstack**, **pickup**, and **putdown** blocks:

**pickup(?ob)**
*clear(?ob) on-table(?ob) arm-empty()* $\longrightarrow$ *ADD holding(?ob)*
                                                    *DEL clear(?ob) on-table(?ob) arm-empty().*

**putdown(?ob)**
*holding(?ob)* $\longrightarrow$ *ADD clear(?ob) arm-empty() on-table(?ob)*
              *DEL holding(?ob).*

**stack(?ob,?underob)**
*clear(?underob) holding(?ob)* $\longrightarrow$ *ADD arm-empty() clear(?ob) on(?ob,?underob)*
                                                *DEL clear(?underob) holding(?ob).*

**unstack(?ob,?underob)**
*on(?ob,?underob) clear(?ob) arm-empty()* $\longrightarrow$ *ADD holding(?ob) clear(?underob)*
                                                            *DEL on(?ob,?underob) clear(?ob) arm-empty().*

Given the simple task of stacking three blocks:

**initial state:** on-table(a) on-table(b) on-table(c)
**goal state:**    on(a,b) on(b,c)

is there a reasonable ordering between the two atomic goals? Intuitively, the *blocks world* domain possesses a very natural goal ordering, namely that the planner should start building each tower from the bottom to the top and not the other way round.[2]

Let us first investigate whether the relation $on(a,b) \leq_e on(b,c)$ holds. Vividly speaking, it asks whether it is still possible to stack the block $a$ on $b$ after $on(b,c)$ has been achieved. As a first step, we run GRAPHPLAN to find out which atoms are exclusive of $on(b,c)$ when the planning graph, which corresponds to this problem, has leveled off. The result is

$$\mathsf{F}_{GP}^{on(b,c)} = \{clear(c),\ on\text{-}table(b),\ holding(c),\ holding(b),\ on(a,c),\ on(c,b),\ on(b,a)\}$$

One observes immediately that these atoms can never be true in a state that satisfies $on(b,c)$.

Secondly, we remove all ground actions which delete $on(b,c)$ (in this case, only the action **unstack(b,c)** satisfies this condition) and obtain the reduced action set $\mathcal{O}_{on(b,c)}$.

Now we are ready to test if $on(a,b) \leq_e on(b,c)$ holds. The only action, which can add $on(a,b)$ is **stack(a,b)**. It has the preconditions $holding(a)$ and $clear(b)$, neither of which is a member of $\mathsf{F}_{GP}^{on(b,c)}$. The test fails and we get $on(a,b) \not\leq_e on(b,c)$.

As a next step, we test whether $on(b,c) \leq_e on(a,b)$ holds. GRAPHPLAN returns the following False set:

$$\mathsf{F}_{GP}^{on(a,b)} = \{clear(b),\ on\text{-}table(a),\ holding(b),\ holding(a),\ on(a,c),\ on(c,b),\ on(b,a)\}$$

The action **unstack(a,b)** is not contained in $\mathcal{O}_{on(a,b)}$ because it deletes $on(a,b)$. The only action which adds $on(b,c)$ is **stack(b,c)**. It needs the preconditions $clear(c)$ and $holding(b)$. The second precondition $holding(b)$ is contained in the set of false facts, i.e., $holding(b) \in \mathsf{F}_{GP}^{on(a,b)}$ and thus, we conclude $on(b,c) \leq_e on(a,b)$. Altogether, we have $on(a,b) \not\leq_e on(b,c)$ and $on(b,c) \leq_e on(a,b)$, which correctly reflects the intuition that $b$ needs to be stacked onto $c$ *before* $a$ can be stacked onto $b$.

Although $\leq_e$ appears to impose very strict conditions on a domain in order to derive a reasonable goal ordering, it succeeds in finding reasonable goal orderings in all available test domains in which such orderings exists. For example, in the *tyreworld*, in *bulldozer* problems, in the *shopping problem* (Russel & Norvig, 1995), the *fridgeworld*, the *glass* domain, the *tower of hanoi* domain, the *link-world*, and the *woodshop*. Its only disadvantage are the computational resources it requires, since building planning graphs, while being theoretically polynomial, is a quite time- and memory-consuming thing to do.[3]

Therefore, the next section presents a fast heuristic computation of goal orderings, which analyzes the domain actions directly and does not need to build planning graphs anymore.

---

2. Note that the goals do not specify where the block $c$ has to go, but leave this to the planner.
3. More recent implementations of planning graphs, which are for example developed for STAN (Fox & Long, 1999) and IPP 4.0 (Koehler, 1999) do not build the graphs explicitly anymore and are orders of magnitude faster than the original GRAPHPLAN implementation, but still the computation of the planning graph takes almost all the time that is needed to determine the $\leq_e$ relations.

### 3.2 Reasonable Goal Orderings derived by a Fast Heuristic Method

One can analyze the available actions directly using a method we will call *Direct Analysis* (DA). It determines an initial value for $\mathsf{F}$ by computing the intersection of all delete lists of all actions which contain $A$ in their add list, as defined in the following equation.

$$\mathsf{F}_{DA}^A := \bigcap_{o \in \mathcal{O},\ A \in add(o)} del(o) \qquad (3)$$

The atoms in this set are all FALSE in a state where $A$ has just been achieved: they are deleted from the state description independently of the action that is used to add $A$. As a short example, let us consider the two actions

$$\longrightarrow \ \text{ADD} \ \{A\} \quad \text{DEL} \ \{C, D\}$$

$$\longrightarrow \ \text{ADD} \ \{A, C\} \quad \text{DEL} \ \{D\}$$

Only the atom $D$ is deleted by both actions, and thus $D$ is the only element initially contained in $\mathsf{F}_{DA}^A$.

However, Equation (3) only says that when $A$ is added then the atoms from $\mathsf{F}_{DA}^A$ will be deleted. It does not say anything about whether it might be possible to reestablish atoms in $\mathsf{F}_{DA}^A$. One can easily imagine that actions exist, which leave $A$ true, and at the same time add such atoms. If this is the case, there are reachable states in which $A$ and atoms from $\mathsf{F}_{DA}^A$ hold.

Now, our goal is to derive an ordering relation that can be easily computed, and that ideally, like the $\leq_e$ relation, is sufficient for the $\leq_r$ relation. Therefore, we want to make sure that the atoms in $\mathsf{F}_{DA}^A$ are really FALSE in any state after $A$ has been achieved. We arrive at an approximation of atoms that remain FALSE by performing a fixpoint reduction on the $\mathsf{F}_{DA}^A$ set, removing those atoms that are *achievable* in the following sense.

**Definition 11 (Achievable Atoms)** *An atom $p$ is* achievable *from a state $s$ given an action set $\mathcal{O}$ (written $\mathcal{A}(s, p, \mathcal{O})$) if and only if*

$$p \in s \ \lor \ \exists\, o \in \mathcal{O} : p \in add(o) \ \land \ \forall\, p' \in pre(o) : \ \mathcal{A}(s, p', \mathcal{O})$$

The definition says that an atom $p$ is achievable from a state $s$ if it holds in $s$, or if there exists an action in the domain, which adds $p$ and whose preconditions are all achievable from $s$. This is a necessary condition for the existence of a plan $\mathcal{P}^{\mathcal{O}}$ from $s$ to a state where $p$ holds.

**Lemma 2** $\exists\, \mathcal{P}^{\mathcal{O}} : \ p \in Result(s, \mathcal{P}^{\mathcal{O}}) \Rightarrow \mathcal{A}(s, p, \mathcal{O})$

**Proof:** The atom $p$ must either already be contained in the state $s$, or it has to be added by a step $o$ out of $\mathcal{P}^{\mathcal{O}}$. In the second case, all preconditions of $o$ need to be established by $\mathcal{P}^{\mathcal{O}}$ in the same way. Thus $p$ and all preconditions of the step, which adds it, are achievable in the sense of Definition 11. ∎

There are two obvious difficulties with Definition 11: First, $p \in s$ must be tested. With complete knowledge about the state $s$, this should not cause any problems. In our case, however, we only have the *generic* state $s_{(A, \neg B)}$ and cannot decide whether an arbitrary atom is contained in it or not. Secondly, we observe an infinite regression over preconditions, which must be tested for achievability.

As for the first problem, it turns out that it is a good heuristic to simply assume $p \notin s$, i.e., no test is performed at all. As for the second problem, in order to avoid infinite looping of the "achievable"-test, one needs to terminate the regression over preconditions at a particular level. The point in question is how far to regress? A quick approximation simply decides "achievable" after the first recursive call.

**Definition 12 (Possibly Achievable Atoms)** *An atom $p$ is* possibly achievable *given an action set $\mathcal{O}$ (written $p\mathcal{A}(p, \mathcal{O})$) if and only if*

$$\exists\, o \in \mathcal{O} : p \in add(o) \ \wedge\ \forall\, p' \in pre(o) :\ \exists\, o' \in \mathcal{O} : p' \in add(o')$$

*holds, i.e., there is an action that adds $p$ and all of its preconditions are add effects of other actions in $\mathcal{O}$.*

If the assumption is justified that none of the atoms $p$ is contained in the state $s$, then being possibly achievable is a necessary condition for being achievable.

**Lemma 3** *Let $s$ be a state for which $p \notin s$ and also $\forall o \in \mathcal{O} : p \in add(o) \Rightarrow pre(o) \cap s = \emptyset$ holds. Then we have*

$$\mathcal{A}(s, p, \mathcal{O}) \Rightarrow p\mathcal{A}(p, \mathcal{O})$$

**Proof:** From $\mathcal{A}(s, p, \mathcal{O})$ and $p \notin s$, we know that there is a step $o \in \mathcal{O}$, $p \in add(o)$, with $\forall\, p' \in pre(o)\ \mathcal{A}(s, p', \mathcal{O})$. We also know that $pre(o) \cap s = \emptyset$, so for each $p' \in pre(o)$ there must be an achiever $o' \in \mathcal{O} : p' \in add(o')$. ∎

The condition that all of the facts $p$ must not be contained in the state $s$ seems to be rather rigid. Nevertheless, the condition of being possibly achievable delivers good results on all of the benchmark domains and it is easy to decide. We can now use this test to both

- perform a fixpoint reduction on the set $\mathsf{F}_{DA}^A$ and
- decide whether an atomic goal $B$ should be ordered before $A$.

The fixpoint reduction, as depicted in Figure 1 below, uses the approximative test $p\mathcal{A}(f, \mathcal{O}^*)$ to remove facts from $\mathsf{F}_{DA}^A$ that can be achieved. It finds *all* these facts under certain restrictions, see below. As a side effect of the fixpoint algorithm, we obtain the set $\mathcal{O}^*$ of actions that our method assumes to be applicable after a state $s_{(A, \neg B)}$. We then order $B$ before $A$ iff it cannot possibly be achieved using these actions.

$$\begin{aligned}
&\mathsf{F}^* := \mathsf{F}^A_{DA} \\
&\mathcal{O}^* := \mathcal{O}_A \setminus \{o \mid \mathsf{F}^* \cap pre(o) \neq \emptyset\} \\
&fixpoint\_reached := \text{FALSE} \\
&\textbf{while } \neg fixpoint\_reached \\
&\qquad fixpoint\_reached := \text{TRUE} \\
&\qquad \textbf{for } f \in \mathsf{F}^* \\
&\qquad\qquad \textbf{if } p\mathcal{A}(f, \mathcal{O}^*) \textbf{ then} \\
&\qquad\qquad\qquad \mathsf{F}^* := \mathsf{F}^* \setminus \{f\} \\
&\qquad\qquad\qquad \mathcal{O}^* := \mathcal{O}_A \setminus \{o \mid \mathsf{F}^* \cap pre(o) \neq \emptyset\} \\
&\qquad\qquad\qquad fixpoint\_reached := \text{FALSE} \\
&\qquad\qquad \textbf{endif} \\
&\qquad \textbf{endfor} \\
&\textbf{endwhile} \\
&\textbf{return } \mathsf{F}^*, \mathcal{O}^*
\end{aligned}$$

Figure 1: Quick, heuristic fixpoint reduction of the set $\mathsf{F}^A_{DA}$.

The computation checks whether atoms of $\mathsf{F}^*$, which is initially set to $\mathsf{F}^A_{DA}$, are possibly achievable using only those actions, which do not delete $A$ and which do not require atoms from $\mathsf{F}^*$ as a precondition. Achievable atoms are removed from $\mathsf{F}^*$, and $\mathcal{O}^*$ gets updated accordingly. If in one iteration, $\mathsf{F}^*$ does not change, the fixpoint is reached, i.e., $\mathsf{F}^*$ will not further decrease and $\mathcal{O}^*$ will not further increase—the final sets $\mathsf{F}^*$ of false facts and $\mathcal{O}^*$ of applicable actions are returned.

Let us illustrate the fixpoint computation with a short example consisting of the empty initial state, the goals $\{A, B\}$, and the following set of actions

| | | | | |
|---|---|---|---|---|
| **op1**: | | $\longrightarrow$ | $ADD \{ A \}$ | $DEL \{ C, D \}$ |
| **op2**: | | $\longrightarrow$ | $ADD \{A, C\}$ | $DEL \{ D \}$ |
| **op3**: | $\{ C \}$ | $\longrightarrow$ | $ADD \{ D \}$ | |
| **op4**: | $\{ D \}$ | $\longrightarrow$ | $ADD \{ B \}$ | |

When assuming that $A$ has been achieved, we obtain $\mathsf{F}^* = \mathsf{F}^A_{DA} = \{D\}$ as the initial value of the False set, since $D$ is the only atom that **op1** and **op2** delete when adding $A$. Figure 2 illustrates a hypothetical planning process. Starting in the empty initial state and trying to achieve $A$ first, we get two different states $s_{(A, \neg B)}$ in which $A$ holds. The atom $D$ does not hold in any of them and thus in both states, no action is applicable that requires $D$ as a precondition. This excludes **op4** from $\mathcal{O}_A$, yielding the initial action set $\mathcal{O}^* = \{\textbf{op1}, \textbf{op2}, \textbf{op3}\}$. Now, **op4** is the only action that can add $B$. Therefore, if we used this action set to see if $B$ can still be achieved, we would find that this is *not* the case. Consequently, without performing the fixpoint computation, we would order $B$ before $A$. But as can be seen in Figure 2, this would not be a reasonable ordering: there is the plan $\langle \textbf{op3}, \textbf{op4} \rangle$ that achieves $B$ from the state $s_{(A, \neg B)} = Result(\mathcal{I}, \textbf{op2})$ without destroying $A$.

The fixpoint computation works us around this problem as follows: There is the action **op3**, which can add the precondition $D$ of **op4** without deleting $A$. When checking $p\mathcal{A}(D, \mathcal{O}^*)$ in the first iteration, the fixpoint procedure finds this action. It then checks

351

whether the preconditions of **op3** are achievable in the sense that they are added by another action. This is the case since the only precondition $C$ is added by **op2**. Thus, $D$ is removed from $\mathsf{F}^*$, which becomes empty now. The action **op4** is put back into the set $\mathcal{O}^*$, which now becomes identical with the action set $\mathcal{O}_A$. This set, in turn, is identical with the original action set $\mathcal{O}$ as no action deletes $A$. The fixpoint process terminates and $B$ will not be ordered before $A$ as it can be achieved using the action **op4**. This correctly reflects the fact that there exists a plan from the state $s_{(A,\neg B)} = Result(\mathcal{I}, \langle \mathbf{op2} \rangle) = \{C, A\}$ to a state that satisfies $B$ without destroying $A$.

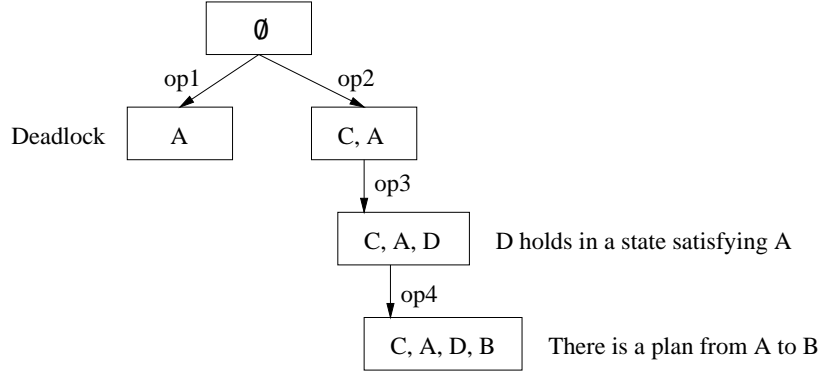

Figure 2: An example illustrating why we need the fixpoint computation.

As already pointed out, the intention behind the fixpoint procedure is the following: Starting from a state $s_{(A,\neg B)}$, we want to know which facts can become TRUE without destroying $A$, and consequently, which actions can become applicable. In the first step, only actions that do not use any of the facts in $\mathsf{F}^A_{DA}$ are applicable, as all those facts are deleted from the state description when $A$ is added. However, such actions may make facts in $\mathsf{F}^A_{DA}$ TRUE, so we want to remove those facts from $\mathsf{F}^A_{DA}$. If we manage to find *all* the facts that can be made TRUE without destroying $A$, then the final set $\mathsf{F}^*$ will contain only those facts that do not hold in a state reachable from $s_{(A,\neg B)}$ without destroying $A$. In this case, the final action set $\mathcal{O}^*$ will contain *all* the actions that can be applied after $s_{(A,\neg B)}$, and we can safely use this action set to determine whether another goal $B$ can still be achieved or not.

However, as we only use the approximative test $p\mathcal{A}(f, \mathcal{O}^*)$ with $f \in \mathsf{F}^*$ to find out if a fact in the current $\mathsf{F}^*$ set is achievable, there may be facts which are achievable without destroying $A$, but which remain in the set $\mathsf{F}^*$. This could exclude actions from the set $\mathcal{O}^*$ which can be safely applied after $s_{(A,\neg B)}$. Under certain restrictions, however, we can prove that this will not happen. In order to do so, we need to impose a restriction on the particular state $s_{(A,\neg B)}$, in which we achieved the goal $A$: If none of the preconditions of actions, which add facts contained in $F^A_{DA}$, occur in the state $s_{(A,\neg B)}$, then the fixpoint procedure will remove all facts from $F^A_{DA}$ that are achievable without destroying $A$. We will use this property of the fixpoint procedure later to show that our heuristic ordering relation approximates reasonable orderings.

**Lemma 4** *Let $(\mathcal{O}, \mathcal{I}, \mathcal{G})$ be a planning problem, and let $A \in \mathcal{G}$ be an atomic goal. Let $s_{(A,\neg B)}$ be a reachable state where $A$ has just been achieved. Let $\mathcal{P}^{\mathcal{O}_A} = \langle o_1, \ldots, o_n \rangle$ be a sequence of actions not destroying $A$. Let $\mathsf{F}^*$ be the set of facts that is returned by the fixpoint computation depicted in Figure 1. If we have*

$$\forall f \in \mathsf{F}_{DA}^A : \forall o \in \mathcal{O}_A : f \in add(o) \Rightarrow pre(o) \cap s_{(A,\neg B)} = \emptyset \qquad (*)$$

*then no fact in $\mathsf{F}^*$ holds in the state that is reached by applying $\mathcal{P}^{\mathcal{O}_A}$, i.e.,*

$$Result(s_{(A,\neg B)}, \mathcal{P}^{\mathcal{O}_A}) \cap \mathsf{F}^* = \emptyset$$

**Proof:**

Let $\mathsf{F}_j^*$ and $\mathcal{O}_j^*$ denote the state of the fact and action sets, respectively, after $j$ iterations of the algorithm depicted in Figure 1. As $\mathsf{F}^*$ only decreases during the computation, we have $\mathsf{F}^* \subseteq \mathsf{F}_j^*$ for all $j$. Let $s_0, \ldots, s_n$ denote the sequence of states that are encountered when executing $\mathcal{P}^{\mathcal{O}_A} = \langle o_1, \ldots, o_n \rangle$ in $s_{(A,\neg B)}$, i.e., $s_0 = s_{(A,\neg B)}$ and $s_i = Result(s_{i-1}, \langle o_i \rangle)$ for $0 \leq i \leq n$. We can assume that each action $o_i$ is applicable in state $s_{i-1}$, i.e., $pre(o_i) \subseteq s_{i-1}$. Otherwise, $o_i$ does not cause any state transition, and we can skip it from $\mathcal{P}^{\mathcal{O}_A}$. Obviously, we have $s_n = Result(s_{(A,\neg B)}, \mathcal{P}^{\mathcal{O}_A})$, so we need to show that $s_n \cap \mathsf{F}^* = \emptyset$. The proof proceeds by induction over the length $n$ of $\mathcal{P}^{\mathcal{O}_A}$.

<u>$n = 0$</u>: $\mathcal{P}^{\mathcal{O}_A} = \langle \rangle$ and $s_n = s_0 = s_{(A,\neg B)}$. All facts in $\mathsf{F}_{DA}^A$ are deleted from the state description when $A$ is added, so we have $s_n \cap \mathsf{F}_{DA}^A = \emptyset$. As $\mathsf{F}_{DA}^A = \mathsf{F}_0^*$ and $\mathsf{F}^* \subseteq \mathsf{F}_0^*$, the proposition follows immediately.

<u>$n \rightarrow n+1$</u>: $\mathcal{P}^{\mathcal{O}_A} = \langle o_1, \ldots, o_n, o_{n+1} \rangle$. From the induction hypothesis, we know that $s_i \cap \mathsf{F}^* = \emptyset$ for $0 \leq i \leq n$. What we need to show is $s_{n+1} \cap \mathsf{F}^* = \emptyset$.

Let $j$ be the step in the fixpoint iteration where $\mathsf{F}_j^* \cap \bigcup_{i=0,\ldots,n} s_i$ becomes empty, i.e., $j$ denotes the iteration in which the intersection of all the states $s_i, i \leq n$ with $\mathsf{F}_j^*$ is empty for the first time. Such an iteration exists, because all the intersections $s_i \cap \mathsf{F}^*$ with $i \leq n$ are empty.

Now each action $o_i, 1 \leq i \leq n+1$ is applicable in state $s_{i-1}$, i.e., $pre(o_i) \subseteq s_{i-1}$, and thus $pre(o_i) \cap \mathsf{F}_j^* = \emptyset$ for all the actions $o_i$ in $\mathcal{P}^{\mathcal{O}_A}$. Therefore, all these actions are contained in $\mathcal{O}_j^*$, as this set contains all the actions out of $\mathcal{O}_A$ whose intersection with $\mathsf{F}_j^*$ is empty. Let us focus on the facts in the state $s_{n+1}$. All these facts are achieved by executing $\mathcal{P}^{\mathcal{O}_A}$ in $s_{(A,\neg B)}$. In other words, there is a plan from $s_{(A,\neg B)}$ to each of these facts. As we have just seen, this plan consists out of actions in $\mathcal{O}_j^*$. Applying Lemma 2 to all the facts $p \in s_{n+1}$ using $s_{(A,\neg B)}$ and $\mathcal{P}^{\mathcal{O}_A}$ $(= \mathcal{P}^{\mathcal{O}_j^*})$, we know that all facts $p$ are achievable using actions from $\mathcal{O}_j^*$.

$$\forall p \in s_{n+1} : \mathcal{A}(s_{(A,\neg B)}, p, \mathcal{O}_j^*)$$

We will now show that those facts $f \in s_{n+1}$ we are interested in, namely the $\mathsf{F}$ facts that are added by $o_{n+1}$ and that are still contained in $\mathsf{F}_j$, are also *possibly* achievable using actions from $\mathcal{O}_j^*$. Let $f$ be a fact $f \in s_{n+1}, f \in \mathsf{F}_j^*$. We apply Lemma 3 using $s_{(A,\neg B)}, f$, and

353

$\mathcal{O}_j^*$. We *can* apply Lemma 3 as obviously $f \notin s_{(A,\neg B)}$, and as $\forall o \in \mathcal{O}_j^* : f \in add(o) \Rightarrow pre(o) \cap s_{(A,\neg B)} = \emptyset$ by prerequisite $(*)$. With $\mathcal{A}(s_{(A,\neg B)}, p, \mathcal{O}_j^*)$, we arrive at

$$\forall f \in s_{n+1} \cap \mathsf{F}_j^* : p\mathcal{A}(f, \mathcal{O}_j^*)$$

What remains to be proven is that all these facts $f$ will be removed from $\mathsf{F}^*$ during the fixpoint computation. With the argumentation above, it is sufficient to show that all the facts $f \in s_{n+1} \cap \mathsf{F}_j^*$ will get tested for $p\mathcal{A}(f, \mathcal{O}_j^*)$ in iteration $j+1$ of the fixpoint computation. These tests will succeed and lead to $s_{n+1} \cap \mathsf{F}_{j+1}^* = \emptyset$, yielding, as desired, $s_{n+1} \cap \mathsf{F}^* = \emptyset$. Remember that $\mathsf{F}_{j+1}^* \supseteq \mathsf{F}^*$. There are two cases, which we need to consider:

1. $j = 0$: all intersections $s_i \cap \mathsf{F}_0^*$ are initially empty, i.e., $s_i \cap \mathsf{F}_{DA}^A = \emptyset$ for $0 \leq i \leq n$. In this case, all facts $f \in s_{n+1} \cap \mathsf{F}_{DA}^A$ are tested for $p\mathcal{A}(f, \mathcal{O}_0^*)$ in iteration $j + 1 = 1$ of the fixpoint computation.

2. $j > 0$: in this case, at least one of the intersections $s_i \cap \mathsf{F}_j^*$ became empty in iteration $j$ by definition of $j$, i.e., at least one fact was removed from $\mathsf{F}^*$ in this iteration. Therefore, the fixpoint has not been reached yet, and the computation performs at least one more iteration, namely iteration $j + 1$. All facts in $\mathsf{F}_j^*$ will be tested in this iteration, in particular all facts $f \in s_{n+1} \cap \mathsf{F}_j^*$.

With these observations, the induction is complete and the proposition is proven. ∎

As has already been said, we now simply order $B$ before $A$, if it is not possibly achievable using the action set that resulted from the fixpoint computation. The ordering relation $\leq_h$ (where $h$ stands for "heuristic") obtained in this way approximates the reasonable goal ordering $\leq_r$.

**Definition 13 (Heuristic Ordering $\leq_h$)** *Let $(\mathcal{O}, \mathcal{I}, \mathcal{G} \supseteq \{A, B\})$ be a planning problem. Let $\mathcal{O}^*$ be the set of actions that is obtained from $\mathcal{O}$ by performing the fixpoint computation shown in Figure 1.*
*The ordering $B \leq_h A$ holds if and only if*

$$\neg p\mathcal{A}(B, \mathcal{O}^*)$$

If $A$ has been reached in a particular state $s_{(A,\neg B)}$ where the assumptions made by the fixpoint computation and by the test for $p\mathcal{A}(B, \mathcal{O}^*)$ are justified, then being not possibly achievable is a sufficient condition for the non-existence of a plan to $B$ that does not temporarily destroy $A$.

**Theorem 5** *Let $(\mathcal{O}, \mathcal{I}, \mathcal{G})$ be a planning problem, and let $A, B \in \mathcal{G}$ be two atomic goals. Let $s_{(A,\neg B)}$ be a reachable state where $A$ has just been achieved, but $B$ is still false, i.e., $B \notin s_{(A,\neg B)}$. Let $\mathsf{F}^*$ and $\mathcal{O}^*$ be the sets of facts and actions, respectively, that are derived by the fixpoint computation shown in Figure 1. If we have*

$$\forall f \in \mathsf{F}_{DA}^A \cup \{B\} : \forall o \in \mathcal{O}_A : f \in add(o) \Rightarrow pre(o) \cap s_{(A,\neg B)} = \emptyset \qquad (**)$$

*then we have*

$$\neg p\mathcal{A}(B, \mathcal{O}^*) \Rightarrow \neg \exists \mathcal{P}^{\mathcal{O}_A} : B \in Result(s_{(A,\neg B)}, \mathcal{P}^{\mathcal{O}_A})$$

**Proof:** Assume that there is a plan $\mathcal{P}^{\mathcal{O}_A} = \langle o_1, \ldots, o_n \rangle$ that does not destroy $A$, but achieves $B$, i.e., $B \in Result(s_{(A, \neg B)}, \langle o_1, \ldots, o_n \rangle)$. With the restriction of $(**)$ to the facts in $\mathsf{F}_{DA}^{A}$, Lemma 4 can be applied to each action sequence $\langle o_1, \ldots, o_{i-1} \rangle$ yielding $Result(s_{(A, \neg B)}, \langle o_1, \ldots, o_{i-1} \rangle) \cap \mathsf{F}^* = \emptyset$. Consequently, each $o_i$ is either

- not applicable in $Result(s_{(A, \neg B)}, \langle o_1, \ldots, o_{i-1} \rangle)$,

- or its preconditions are contained in $Result(s_{(A, \neg B)}, \langle o_1, \ldots, o_{i-1} \rangle)$, yielding $pre(o_i) \cap \mathsf{F}^* = \emptyset$.

In the first case, we simply skip $o_i$ as it does not have any effects. In the second case, $o_i \in O^*$ follows. Thus, we have a plan constructed out of actions in $O^*$ that achieves $B$ from $s_{(A, \neg B)}$. Applying Lemma 2 leads us to $\mathcal{A}(s_{(A, \neg B)}, B, O^*)$. We have $B \notin s_{(A, \neg B)}$. We also know, from $(**)$ with respect to $B$, as $\mathcal{O}^* \subseteq \mathcal{O}_A$, that $\forall o \in \mathcal{O}^* : B \in add(o) \Rightarrow pre(o) \cap s_{(A, \neg B)} = \emptyset$ holds. Therefore, we can now apply Lemma 3 and arrive at $p\mathcal{A}(B, \mathcal{O}^*)$, which is a contradiction. ∎

We return to the *blocks world* example and show how the computation of $\leq_h$ proceeds. Let us first investigate whether $on(a, b) \leq_h on(b, c)$ holds. The initial value for $\mathsf{F}_{DA}^{on(b,c)}$ is obtained from the delete list of the **stack(b,c)** action, which is the only one that adds this goal.

$$\mathsf{F}_{DA}^{on(b,c)} = \{clear(c), holding(b)\}$$

Intuitively, it is immediately clear that neither of these facts can ever hold in a state where $on(b, c)$ is true: if $b$ is on $c$, then $c$ is not clear and the gripper cannot hold $b$. It turns out that the fixpoint computation respects this intuition and leaves the set $\mathsf{F}_{DA}^{on(b,c)}$ unchanged, yielding $\mathsf{F}^* = \{clear(c), holding(b)\}$. We do not repeat the fixpoint process in detail here, because it can be reconstructed from Figure 1 and the details are not necessary for understanding how the correct ordering relations are derived. In short, for both facts there are achievers in the reduced action set, but all of them need preconditions for which no achiever is available. For example, $holding(b)$ can be achieved by either an **unstack** or a **pickup** action. Both either need $b$ to stand on another block or to stand on the table. All actions that can achieve these facts need $holding(b)$ to be TRUE and are thus excluded from the reduced action set.

After finishing the fixpoint computation, the planner tests $p\mathcal{A}(on(a, b), \mathcal{O}^*)$, where $\mathcal{O}^*$ contains all actions except those that delete $on(b, c)$ and those that use $clear(c)$ or $holding(b)$ as a precondition. It finds that the action **stack(a,b)** adds $on(a, b)$. The preconditions of this action are $holding(a)$ and $clear(b)$. These conditions are added by the actions **pickup(a)** and **unstack(a,b)**, respectively, which are both contained in $\mathcal{O}^*$: neither of them needs $c$ to be clear or $b$ to be in the gripper. Thus, the test finds that in fact, $on(a, b)$ is possibly achievable using the actions in $\mathcal{O}^*$, and no ordering is derived, i.e., $on(a, b) \not\leq_h on(b, c)$ follows.

Now, the other way round, $on(b, c) \leq_h on(a, b)$ is tested. The initial value for $\mathsf{F}_{DA}^{on(a,b)}$ is obtained from the single action **stack(a,b)** as

$$\mathsf{F}_{DA}^{on(a,b)} = \{clear(b), holding(a)\}$$

Again, the fixpoint computation does not cause any changes, resulting in $\mathsf{F}^* = \{clear(b), holding(a)\}$. The process now tests whether $p\mathcal{A}(on(b,c), \mathcal{O}^*)$ holds, where $\mathcal{O}^*$ contains all actions except those that delete $on(a,b)$ and those that use $clear(b)$ or $holding(a)$ as a precondition. The only action that can add $on(b,c)$ is **stack(b,c)**. This action needs as preconditions the facts $holding(b)$ and $clear(c)$. The process now finds that a crucial condition for achieving the first fact is violated: Each action that can achieve $holding(b)$ has $clear(b)$ as a precondition, because $b$ must be clear first before the gripper can hold it. Since $clear(b)$ is an element of $\mathsf{F}^*$, none of the actions achieving $holding(b)$ is contained in $\mathcal{O}^*$. Consequently, the test for $p\mathcal{A}(on(b,c), \mathcal{O}^*)$ fails and we obtain the ordering $on(b,c) \leq_h on(a,b)$. This makes sense as the gripper cannot grasp $b$ and stack it onto $c$ anymore, once $on(a,b)$ is achieved.

### 3.3 On Forced Goal Orderings and Invertible Planning Problems

So far, we have introduced two easily computable ordering relations $\leq_h$ and $\leq_e$ that both approximate the reasonable goal ordering $\leq_r$. One might wonder why we do not invest any effort in trying to find *forced* goal orderings. There are two reasons for that:

1. As we have already seen in Section 2, any forced goal ordering is also a reasonable goal ordering, i.e., a method that approximates the latter can also be used as a crude approximation to the former.

2. Many benchmark planning problems are *invertible* in a certain sense. Those problems do not contain forced orderings anyway.

In this section, we elaborate in detail the second argument. The results are a bit more general than necessary at this point. We want to make use of them later when we show that the Agenda-Driven planning algorithm we propose is complete with respect to a certain class of planning problems. We proceed by formally defining this class of planning problems, show that these problems do not contain forced orderings, and identify a sufficient criterion for the membership of a problem in this class. Finally, we demonstrate that many benchmark planning problems do in fact satisfy this criterion. For a start, we introduce the notion of a *deadlock* in a planning problem.

**Definition 14 (Deadlock)** *Let $(\mathcal{O}, \mathcal{I}, \mathcal{G})$ be a planning problem. A reachable state $s$ is called a* deadlock *iff there is no sequence of actions that leads from $s$ to the goal, i.e., iff $s = Result(\mathcal{I}, \mathcal{P}^\mathcal{O})$ and $\neg \exists \, \mathcal{P}'^\mathcal{O} : \mathcal{G} \subseteq Result(s, \mathcal{P}'^\mathcal{O})$.*

The class of planning problems we are interested in is the class of problems that are *deadlock-free*. Naturally, a problem is called deadlock-free if none of its reachable states is a deadlock in the sense of Definition 14.

Non-trivial forced goal orderings imply the existence of deadlocks (remember that an ordering $B \leq_f A$ or $B \leq_r A$ is called *trivial* iff there is no state $s_{(A, \neg B)}$ at all).

**Lemma 5** *Let $(\mathcal{O}, \mathcal{I}, \mathcal{G})$ be a planning problem, and let $A, B \in \mathcal{G}$ be two atomic goals. If there is a non-trivial forced ordering $B \leq_f A$ between $A$ and $B$, then there exists a deadlock state $s$ in the problem.*

**Proof:** Recalling Definition 9 and assuming non-triviality of $\leq_f$, we know that there is at least one state $s_{(A,\neg B)}$ where $A$ is made TRUE, but $B$ is still FALSE. From Definition 7, we know that there is no plan in any such state that achieves $B$. In particular, it is not possible to achieve all *goals* starting out from $s_{(A,\neg B)}$. Thus, the state $s := s_{(A,\neg B)}$ must be a deadlock. ∎

We will now investigate deadlocks in more detail and discuss that most of the commonly used benchmark problems do not contain them, i.e., they are deadlock-free. With Lemma 5, we then also know that such domains do not contain non-trivial forced goal orderings either—so there is not much point in trying to find them. We do not care about trivial goal orderings. Such orderings force any reasonable planning algorithm to consider the goals in the correct order.

The existence of deadlocks depends on structural properties of a planning problem: There must be action sequences, which, once executed, lead into states from which the goals cannot be reached anymore. These sequences must have undesired effects, which cannot be inverted by any other sequence of actions in $\mathcal{O}$. Changing perspective, one obtains a hint on how a sufficient condition for the *non-existence* of deadlocks might be defined. Assume we have a planning problem where the effects of each action sequence in the domain can be inverted by executing a certain other sequence of actions. In such an *invertible* planning problem, it is in particular possible to get back to the *initial state* from each reachable state. Therefore, if such a problem is solvable, then it does not contain deadlocks: From any state, one can reach all goals by going back to the initial state first, and then execute an arbitrary solution thereafter. We will now formally define the notion of invertible planning problems, and turn the above argumentation into a proof.

**Definition 15 (Invertible Planning Problem)** *Let $(\mathcal{O}, \mathcal{I}, \mathcal{G})$ be a planning problem, and let $s$ denote the states that are reachable from $\mathcal{I}$ with actions from $\mathcal{O}$. The problem is called* invertible *if and only if*

$$\forall s: \ \forall \mathcal{P}^{\mathcal{O}}: \ \exists \overline{\mathcal{P}}^{\mathcal{O}}: \ Result(Result(s, \mathcal{P}^{\mathcal{O}}), \overline{\mathcal{P}}^{\mathcal{O}}) = s$$

**Theorem 6** *Let $(\mathcal{O}, \mathcal{I}, \mathcal{G})$ be an invertible planning problem, for which a solution exists. Then $(\mathcal{O}, \mathcal{I}, \mathcal{G})$ does not contain any deadlocks.*

**Proof:** Let $s = Result(\mathcal{I}, \mathcal{P}_s^{\mathcal{O}})$ be an arbitrary reachable state. As the problem is invertible, we know that there is a sequence of actions $\overline{\mathcal{P}}_s^{\mathcal{O}}$ for which $Result(s, \overline{\mathcal{P}}_s^{\mathcal{O}}) = \mathcal{I}$ holds. As the problem is solvable, we have a solution plan $\mathcal{P}^{\mathcal{O}}$ starting from $\mathcal{I}$ and achieving $\mathcal{G} \subseteq Result(\mathcal{I}, \mathcal{P}^{\mathcal{O}})$. Together, we obtain $\mathcal{G} \subseteq Result(Result(s, \overline{\mathcal{P}}_s^{\mathcal{O}}), \mathcal{P}^{\mathcal{O}})$. Therefore, the concatenation of $\overline{\mathcal{P}}_s^{\mathcal{O}}$ and $\mathcal{P}^{\mathcal{O}}$ is a solution plan executable in $s$ and consequently, $s$ is no deadlock. ∎

We now know that invertible planning problems, if solvable, do not contain deadlocks and consequently, they do not contain (non-trivial) forced goal orderings. What we will see next is that, as a matter of fact, most benchmark planning problems *are* invertible. We arrive at a sufficient condition for invertibility through the notion of *inverse actions*.

**Definition 16 (Inverse Action)** *Given an action set $\mathcal{O}$ containing an action $o$ of the form $pre(o) \longrightarrow add(o)\ del(o)$. An action $\overline{o} \in \mathcal{O}$ is called* inverse *to $o$ if and only if $\overline{o}$ has the form $pre(\overline{o}) \longrightarrow add(\overline{o})\ del(\overline{o})$ and satisfies the following conditions*

1. $pre(\overline{o}) \subseteq pre(o) \cup add(o) \setminus del(o)$
2. $add(\overline{o}) = del(o)$
3. $del(\overline{o}) = add(o)$

Under certain conditions, applying an inverse action leads back to the state one started from.

**Lemma 6** *Let $s$ be a state and $o$ be an action, which is applicable in $s$. If $del(o) \subseteq pre(o)$ and $s \cap add(o) = \emptyset$ hold, then an action $\overline{o}$ that is inverse to $o$ in the sense of Definition 16 is applicable in $Result(s, \langle o \rangle)$ and $Result(Result(s, \langle o \rangle), \langle \overline{o} \rangle) = s$ follows.*

**Proof:** As $o$ is applicable in $s$, we have $pre(o) \subseteq s$. The atoms in $add(o)$ are added, and the atoms in $del(o)$ are removed from $s$, so altogether we have

$$Result(s, \langle o \rangle) \supseteq (pre(o) \cup add(o)) \setminus del(o) \supseteq pre(\overline{o})$$

Thus, $\overline{o}$ is applicable in $Result(s, \langle o \rangle)$.
Furthermore, we have $Result(s, \langle o \rangle) = s \cup add(o) \setminus del(o)$ and with that

$$
\begin{aligned}
&Result(Result(s, \langle o \rangle), \langle \overline{o} \rangle) \\
=\ & Result(s \cup add(o) \setminus del(o), \langle \overline{o} \rangle) \\
=\ & (s \cup add(o) \setminus del(o)) \cup add(\overline{o}) \setminus del(\overline{o}) \\
=\ & (s \cup add(o) \setminus del(o)) \cup del(o) \setminus add(o) &&\text{(cf. Definition 16)} \\
=\ & s \cup add(o) \setminus add(o) &&\text{(because } del(o) \subseteq pre(o) \subseteq s\text{)} \\
=\ & s &&\text{(because } s \cap add(o) = \emptyset\text{)}
\end{aligned}
$$

∎

Lemma 6 states two prerequisites: (1) inclusion of the operator's delete list in its preconditions and (2) an empty intersection of the operator's add list with the state where it is applicable. A planning problem is called invertible if it meets both prerequisites and if there is an inverse to each action.

**Theorem 7** *Given a planning problem $(\mathcal{O}, \mathcal{I}, \mathcal{G})$ with the set of ground actions $\mathcal{O}$ satisfying $del(o) \subseteq pre(o)$ and $pre(o) \subseteq s \Rightarrow add(o) \cap s = \emptyset$ for all actions and reachable states $s$. If there is an inverse action $\overline{o} \in \mathcal{O}$ for each action $o \in \mathcal{O}$, then the problem is invertible.*

**Proof:** Let $s$ be a reachable state, and let $\mathcal{P}^{\mathcal{O}} = \langle o_1, \ldots o_n \rangle$ be a sequence of actions. We need to show the existence of a sequence $\overline{\mathcal{P}}^{\mathcal{O}}$ for which

$$Result(Result(s, \mathcal{P}^{\mathcal{O}}), \overline{\mathcal{P}}^{\mathcal{O}}) = s \qquad (***)$$

holds. We define $\overline{\mathcal{P}}^{\mathcal{O}} := \langle \overline{o_n}, \ldots, \overline{o_1} \rangle$, and prove $(***)$ by induction over $n$.

$\underline{n = 0}$: Here, we have $\mathcal{P}^{\mathcal{O}} = \overline{\mathcal{P}}^{\mathcal{O}} = \langle \rangle$, and $Result(Result(s, \langle \rangle), \langle \rangle) = s$ is obvious.

$\underline{n \to n+1}$: Now $\mathcal{P}^{\mathcal{O}} = \langle o_1, \ldots, o_n, o_{n+1} \rangle$. From the induction hypothesis we know that $Result(Result(s, \langle o_1, \ldots, o_n \rangle), \langle \overline{o_n}, \ldots, \overline{o_1} \rangle) = s$. To make the following a bit more readable, let $s'$ denote $s' := Result(s, \langle o_1, \ldots, o_n \rangle)$. We have

$$\begin{aligned}
& Result(Result(s, \langle o_1, \ldots, o_{n+1} \rangle), \langle \overline{o_{n+1}}, \ldots, \overline{o_1} \rangle) \\
=\ & Result(Result(s', \langle o_{n+1} \rangle), \langle \overline{o_{n+1}}, \ldots, \overline{o_1} \rangle) \\
=\ & Result(Result(Result(s', \langle o_{n+1} \rangle), \langle \overline{o_{n+1}} \rangle), \langle \overline{o_n}, \ldots, \overline{o_1} \rangle) \\
=\ & Result(s', \langle \overline{o_n}, \ldots, \overline{o_1} \rangle) & \text{(cf. Lemma 6 on } s' \text{ and } o_{n+1}) \\
=\ & s & \text{(per induction)}
\end{aligned}$$

∎

Altogether, we know now that invertible problems, if solvable, do not contain forced orderings. We also know that problems, where there is an inverse action to each action in $\mathcal{O}$, are invertible following Theorem 7. Theorem 7 requires $del(o) \subseteq pre(o)$ to hold for each action $o$, and $pre(o) \subseteq s \Rightarrow add(o) \cap s = \emptyset$ to hold for all actions and reachable states $s$. We will see that all conditions, (a) inclusion of the delete list in the precondition list, (b) empty intersection of an action's add list with reachable states where it is applicable, and (c) existence of inverse actions, hold in most currently used benchmark domains.[4]

Concerning the condition (a) that actions only delete facts they require as preconditions, one finds this phenomenon in *all* domains that are commonly used by the planning community, at least in those that are known to the authors. It is just something that seems to hold in any reasonable logical problem formulation. Some authors even postulate it as an assumption for their algorithms to work, cf. (Fox & Long, 1998).

Similarly in the case of conditions (b) and (c): One usually finds inverse actions in benchmark domains. Also, an action's preconditions usually imply—by state invariants—that its add effects are all FALSE. For example in the *blocks world*, **stack** and **unstack** actions invert each other, and an action's add effects are exclusive of its preconditions—the former are contained in the union of the False constructed for the preconditions, see Section 3.1. Similarly in domains that deal with logistics problems, for example *logistics*, *trains*, *ferry*, *gripper* etc., one can often find inverse pairs of actions with their preconditions always excluding the add effects. Sometimes, two different ground instances of the same operator schema yield an inverse pair. For example, in *gripper*, the two ground instances

**move(roomA, roomB)**
*at-robby(roomA)* $\longrightarrow$ ADD *at-robby(roomB)* DEL *at-robby(roomA)*.

and

---

4. In order to avoid reasoning about reachable states in condition (b), one could also postulate that an action has all of its add effects as *negative preconditions*, cf. (Jonsson, Haslum, & Bäckström, 2000). This is, however, not commonly used in the typical planning benchmark problems.

**move(roomB, roomA)**
*at-robby(roomB)* ⟶ *ADD at-robby(roomA) DEL at-robby(roomB).*

of the **move(?from,?to)** operator schema invert each other. Similarly, in *towers of hanoi*, where there is only the single **move** operator schema, an inverse instance can be found for each ground instance of the schema, and the add effects are always FALSE when the preconditions are TRUE.

Only very rarely, non-invertible actions can be found in benchmark domains. If they occur, their role in the domain is often quite limited as for example the operators **cuss** and **inflate** in Russel's Tyreworld.

**cuss**
⟶ *DEL annoyed().*

**inflate(?x:wheel)**
*have(pump) not-inflated(?x) intact(?x)* ⟶ *ADD inflated(?x) DEL not-inflated(?x).*

Obviously, there is not much point in defining something like a **decuss** or a **deflate** operator. More formally speaking, none of the ground actions to these operators destroys a goal or a precondition of any other action in the domain. Therefore, it does not matter that their effects cannot be inverted. In particular, no forced goal ordering can be derived wrt. these actions. [5]

The importance of inverse actions in real-world domains has also been discussed by Nayak and Williams (1997), who describe the planner BURTON controlling the Cassini spacecraft. In contrast to these domains, problems such as those for example used by Barrett et al. in (1994) almost never contain inverse actions. Consequently, in these domains plenty of forced goal orderings could be discovered and used by a planner to avoid deadlock situations. The widespread, although perhaps unconscious use of invertible problems for benchmarking is a current phenomenon related to STRIPS descending planning systems. As one of the anonymous reviewers pointed out to us, quite a number of non-invertible planning problems have also been proposed in the planning literature, e.g., the *register assignment* problem (Nilsson, 1980), the *robot crossing a road* problem (Sanborn & Hendler, 1988), some instances of manufacturing problems (Regli, Gupta, & Nau, 1995), and the *Yale Shooting* problem (McDermott & Hanks, 1987). For these problems, i.e., for problems that are not invertible, one could—in the spirit of argument 1 at the very beginning of this section—simply use $\leq_e$ and $\leq_h$ to approximate forced orderings if one is interested in finding at least those. More precisely, $\leq_e$ and $\leq_h$ are methods that might detect forced orderings—as those are also reasonable—but that might also find more, not necessarily forced, orderings. If one is not interested in finding *only* the forced orderings, this is a possible way to go. For example, in a simple *blocks world* modification where blocks cannot be unstacked anymore once they are stacked—which *forces* the planner to build the stacks bottom up—both $\leq_e$ and $\leq_h$ are still capable of finding the correct goal orderings.

---

5. The **cuss** operator, by the way, is the only one known to the authors that deletes a fact it is not using as a precondition. It is also the only one we know that could be removed from the domain description without changing anything.

### 3.4 An Extension of Goal Orderings to ADL Actions

The orderings, which have been introduced so far, can be easily extended to deal with ground ADL actions having conditional effects and using negation instead of delete lists. Such actions have the following syntactic structure:

$$o: \begin{aligned} \phi_0(o) &= pre_0(o) \longrightarrow \textit{eff}_0^+(o), \textit{eff}_0^-(o) \\ \phi_1(o) &= pre_1(o) \longrightarrow \textit{eff}_1^+(o), \textit{eff}_1^-(o) \\ &\vdots \\ \phi_n(o) &= pre_n(o) \longrightarrow \textit{eff}_n^+(o), \textit{eff}_n^-(o) \end{aligned}$$

All unconditional elements of the action are summarized in $\phi_0(o)$: The precondition of the action is denoted with $pre_0(o)$, and its unconditional positive and negative effects with $\textit{eff}_0^+(o)$ and $\textit{eff}_0^-(o)$, respectively. Each conditional effect $\phi_i(o)$ consists of an effect condition (antecedent) $pre_i(o)$, and the positive and negative effects $\textit{eff}_i^+(o)$ and $\textit{eff}_i^-(o)$. Additionally, we denote with $\Phi(o)$ the set of all unconditional and conditional effects, i.e., $\Phi(o) = \{\phi_0(o), \phi_1(o), \ldots, \phi_n(o)\}$.

The computation of $\leq_e$ immediately carries over to ADL actions when an extension of planning graphs is used, which can handle conditional effects, e.g., IPP (Koehler, Nebel, Hoffmann, & Dimopoulos, 1997) or SGP (Anderson & Weld, 1998). One simply takes the set of exclusive facts that is returned by these systems to determine the set $\mathsf{F}_{GP}^A$. The test from Definition 10, which decides whether there is an ordering $B \leq_e A$ of two atomic goals $A$ and $B$, is extended to ADL as follows.

**Definition 17 (Ordering $\leq_e$ for ADL)** *Let $(\mathcal{O}, \mathcal{I}, \mathcal{G} \supseteq \{A, B\})$ be a planning problem. Let $\mathsf{F}_{GP}^A$ be the* False *set for $A$. The ordering $B \leq_e A$ holds if and only if*

$$\forall\, o \in \mathcal{O},\ \phi_i(o) \in \Phi(o) : B \in \textit{eff}_i^+(o) \wedge A \notin D_i(o) \ \Rightarrow\ (pre_i(o) \cup pre_0(o)) \cap \mathsf{F}_{GP}^A \neq \emptyset$$

*Here, $D_i(o)$ denotes all negative effects that are implied by the conditions of $\phi_i(o)$.*

$$D_i(o) := \begin{cases} \textit{eff}_0^-(o) \cup \bigcup_{pre_j(o) \subseteq pre_i(o)} \textit{eff}_j^-(o) & i \neq 0 \\ \textit{eff}_0^-(o) & i = 0 \end{cases}$$

Thus, $B$ is ordered before $A$ if all (unconditional or conditional) effects that add $B$ either imply an effect that deletes $A$, or need conditions that cannot be made TRUE together with $A$. Note that an effect $\phi_i$ requires all the conditions in $pre_i(o) \cup pre_0(o)$ to be satisfied, which is impossible in any state where $A$ holds because of the non-empty intersection with $\mathsf{F}_{GP}^A$.

The computation of $\leq_h$ requires a little more adaptation effort. In order to obtain the set $F_{DA}^A$, we now need to investigate the conditional effects as well. For each action that has $A$ as a conditional or unconditional effect, we determine which atoms are negated by it, no matter which effect is used to achieve $A$. We obtain these atoms by intersecting the appropriate sets $D_i(o)$.

$$D(o) := \bigcap_{A \in \textit{eff}_i^+(o)} D_i(o)$$

These are exactly the facts that are always deleted by $o$ when achieving $A$, no matter which effect we use.

The intersection of the sets $D(o)$ for all actions $o$ yields the desired set $F_{DA}^A$. Let us consider the following small example to clarify the computation.

$$\begin{aligned} \phi_0(o) = \{U\} &\longrightarrow \{W\}\ \{\neg X\}; \\ \phi_1(o) = \{V, W\} &\longrightarrow \{A\}\ \{\neg X\}; \\ \phi_2(o) = \{W\} &\longrightarrow \{U\}\ \{\neg Y\} \end{aligned}$$

We obtain $D_1(o) = \{\neg X\} \cup \{\neg Y\} = \{\neg X, \neg Y\}$, because the precondition of $\phi_2(o)$ is implied by the first conditional effect $\phi_1(o)$. As $\phi_1(o)$ is the only effect that can achieve $A$, we get $D(o) = D_1(o) = \{\neg X, \neg Y\}$.

We obtain a smaller set $D(o)$, if we add $A$ as an unconditional positive effect of the action.

$$\begin{aligned} \phi_0(o) = \{U\} &\longrightarrow \{W, A\}\ \{\neg X\}; \\ \phi_1(o) = \{V, W\} &\longrightarrow \{A\}\ \{\neg X\}; \\ \phi_2(o) = \{W\} &\longrightarrow \{U\}\ \{\neg Y\} \end{aligned}$$

In this case, we need to intersect the sets $D_0(o) = \{\neg X\}$ and $D_1(o) = \{\neg X, \neg Y\}$, yielding $D(o) = \{\neg X\}$. This reflects the fact that, when achieving $A$ via the unconditional effect of $o$, only $X$ gets removed from the state.

The fixpoint computation requires to adapt the computation of $\mathcal{O}^*$. First, we repeat the same steps as in the case of simple STRIPS actions and consider the unconditional negative effects and the intersection of the preconditions with the False set:

$$\mathcal{O}^* := \mathcal{O} \setminus \{o \mid A \in \mathit{eff}_0^-(o) \vee \mathsf{F}_{DA}^A \cap pre_0(o) \neq \emptyset\}$$

Then, we additionally remove from each action the conditional effects that either imply the deletion of $A$ or have an impossible effect condition.

$$\mathcal{O}^* := red(\mathcal{O}^*) = \{red(o) | o \in \mathcal{O}^*\}$$

Here, $red$ is a function $red(o) : o \mapsto o'$ such that

$$\Phi(o') = \Phi(o) \setminus \{\phi_k(o) \mid A \in D_k(o) \vee pre_k(o) \cap \mathsf{F}_{DA}^A \neq \emptyset\}$$

Finally, we need to redefine Definition 12, which expresses the conditions under which a fact is believed to be *possibly achievable* given a certain set of operators $\mathcal{O}$.

**Definition 18 (Possibly Achievable Atoms for ADL)** *An atom $p$ is possibly achievable given an action set $\mathcal{O}$ (written $p\mathcal{A}(p, \mathcal{O})$) if and only if*

$$\begin{aligned} \exists\, o \in \mathcal{O},\ \phi_i \in \Phi(o) : p \in \mathit{eff}_i^+(o) \wedge \\ \forall\, p' \in (pre_i(o) \cup pre_0(o)) : \exists\, o' \in \mathcal{O},\ \phi_{i'} \in \Phi(o') : p' \in \mathit{eff}_{i'}^+(o') \end{aligned}$$

*holds, i.e., there is a positive effect for $p$ and all of its conditions and preconditions can be made true by other effects in the reduced action set.*

The process, which decides whether an atomic goal $B$ is heuristically ordered before another goal $A$ (i.e., whether $B \leq_h A$ holds) proceeds in exactly the same way as described in Section 3.2: The False set $\mathsf{F}_{DA}^A$ for $A$ is reduced by the fixpoint computation, which remains unchanged, but employs the updated routines for computing $\mathcal{O}^*$ and for deciding $p\mathcal{A}(f, \mathcal{O}^*)$. As a result, $B$ is ordered before $A$ ($B \leq_h A$) if and only if it is not possibly achievable $p\mathcal{A}(B, \mathcal{O}^*)$ using the action set that results from the fixpoint.

## 4. The Use of Goal Orderings During Planning

After having determined the ordering relations that hold between pairs of atomic goals from a given goal set, the question is how to make use of them during planning. Several proposals have been made in the literature, see Section 6 for a detailed discussion. In this paper, we propose a novel approach that extracts an explicit ordering between subsets of the goal set—called the *goal agenda*. The planner, in our case IPP, is then run successively on the planning subproblems represented in the agenda.

### 4.1 The Goal Agenda

The first step one has to take for computing the goal agenda is to perform a so-called *goal analysis*. During goal analysis, each pair $A, B \in \mathcal{G}$ of atomic goals must be examined in order to find out whether an ordering relation $A \leq B$, or $B \leq A$, or both, or none holds between them. For the ordering relation $\leq$, an arbitrary definition can be used. In our experiments, the relation $\leq$ was always either $\leq_e$ or $\leq_h$.

After having determined all ordering relations that hold between atomic goals, we want to split the goal set into smaller sets based on these relations, and we want to order the smaller sets, also based on these relations. More precisely, our goal is to have a sequence of goal sets $G_1, \ldots, G_n$ with

$$\bigcup_{i=1}^{n} G_i = \mathcal{G}$$

and

$$G_i \cap G_j = \emptyset$$

for $i \neq j, 1 \leq i, j \leq n$. We also want the sequence of goal sets to respect the ordering relations that have been derived between atomic goals. To make this explicit, we first introduce a simple representation for the detected atomic orderings: the *goal graph G*.

$$G := (V, E)$$

where

$$V := \mathcal{G}$$

and

$$E := \{(A, B) \in \mathcal{G} \times \mathcal{G} \mid A \leq B\}$$

Now, the desired properties, which the sequence of goal sets should possess, can be easily stated:

- Goals $A, B$ that lie on a cycle in $G$ belong to the same set, i.e., $A, B \in G_i$.

- If $G$ contains a path from a goal $A$ to a goal $B$, but not vice versa, then $A$ is ordered before $B$, i.e., $A \in G_i$ and $B \in G_j$ with $i < j$.

These are the only properties that appear to be reasonable for a goal-set sequence respecting the atomic orderings. We will now introduce a simple algorithmic method that does produce a sequence of goal sets which meets these requirements.

First of all, the transitive closure of $G$ is computed. This can be done in at most cubic time in the size of the goal set (Warshall, 1962). Then, for each node $A$ in the transitive closure, the ingoing edges $A_{in}$ and outgoing edges $A_{out}$ are counted. All disconnected nodes with $A_{in} = A_{out} = 0$ are moved into a separate set of goals $G$-$sep$ containing now those atomic goals, which do not participate in a $\leq$ relation. For all other nodes $A$, their *degree* $d(A) = A_{in} - A_{out}$ is determined as the difference between the number of ingoing edges and the number of outgoing edges. Nodes with identical degree are merged into one set. The sets are then ordered by increasing degree and yield our desired sequence of goal sets. The only problem remaining is the set $G$-$sep$. If it is non-empty, it is not clear in which place to put it.

Let us consider a small example of the process. Figure 3 depicts on the left the goal graph, which results from the goal set $\mathcal{G} = \{A, B, C, D, E\}$ and the ordering relations $A \leq B, B \leq C$ and $B \leq D$, and its transitive closure on the right.

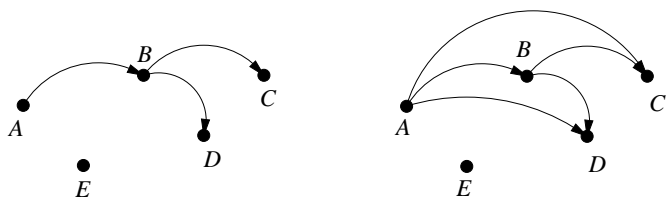

Figure 3: On the left, the goal graph depicting the $\leq$ relations between the atomic subgoals. On the right, the transitive closure of this graph.

In Figure 4, the number of in- and outgoing edges of each goal, the corresponding degrees, and resulting goal-set sequence are shown.

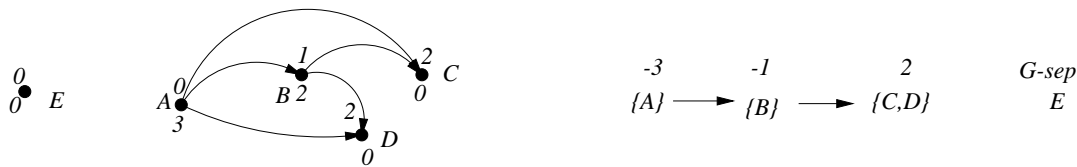

Figure 4: On the left, the number of in- and outgoing edges for each node. On the right, the degree of the nodes and the merged sets of goals having same degree. The node $E$ becomes a member of the $G$-$sep$ set and remains unordered.

It is not difficult to verify that the resulting goal sequence respects the atomic goal orderings:

- Nodes occurring on a cycle in a graph have isomorphic in- and outgoing edges in the transitive closure of that graph. In particular, they have the same degree and get merged into the same set $G_i$.

- Say we have a graph, where there is a path from $A$ to $B$, but not vice versa. Then, in the transitive closure of that graph, we will have an edge from $A$ to each node that $B$ has a path to, and additionally the edge from $A$ to $B$, i.e., $A_{out} > B_{out}$ follows. Similarly, we have an ingoing edge to $B$ for each node that has a path to $A$, and additionally, the edge from $A$ to $B$, which gives us $B_{in} > A_{in}$. Altogether, $d(A) = A_{in} - A_{out} < B_{in} - A_{out} < B_{in} - B_{out} = d(B)$ and thus, the degree of $A$ is smaller than the degree of $B$ and as required, $A$ gets ordered before $B$.

Note that nothing is said in this argumentation about the set of unordered goals, *G-sep*. This set could, in principle, be inserted anywhere in the sequence with the resulting sequence still respecting the atomic orderings. A possible heuristic may use this goal set as the first in the sequence, because apparently there is no problem to reach all other goals after the goals in this set have been achieved. Another heuristic could put this set at the end as there is neither a problem to reach this goal set from all other goals. We have decided to deal with the problem in a more sophisticated way by trying to derive an ordering relation between *G-sep* and the other goal sets $G_i$ that have already been derived. In order to do so, we need to extend our definitions of goal orderings to *sets* of goals.

### 4.2 Extension of Goal Orderings to Goal Sets

Given a set of atomic goals, it has always been a problem which of the exponentially many subsets should be compared with each other in order to derive a reasonable goal ordering between goal sets. A consideration of all possible subsets is out of question, because it will result in an exponential overhead. The partial goal agenda that we have obtained so far offers one possible answer. It suggests taking the set *G-sep* and trying to order it with respect to the goal sets emerging from the goal graph.

Given a planning problem $(\mathcal{O}, I, \mathcal{G})$ and two subsets of atomic goals $\{A_1, \ldots, A_n\} \subseteq \mathcal{G}$ and $\{B_1, \ldots, B_k\} \subseteq \mathcal{G}$, the definition of $\leq_e$ and $\leq_h$ for sets of atomic goals is straightforward. For the sake of simplicity, we consider only STRIPS actions here. The definitions can be directly extended to ADL.

To define an ordering $\leq_E$, which extends $\leq_e$ to sets, we begin by defining a set $F_{GP}^{\{A_1,\ldots,A_n\}}$ of all atoms, which are exclusive of at least one atomic goal $A_i$ in the planning graph generated for $(\mathcal{O}, I, \mathcal{G})$:

$$F_{GP}^{\{A_1,\ldots,A_n\}} := \{p \mid p \text{ is exclusive of at least one } A_i \text{ when the graph has leveled off }\}$$

The set $\mathcal{O}_{\{A_1,\ldots,A_n\}}$ is obtained accordingly by removing from $\mathcal{O}$ all actions that delete at least one of the $A_i$, i.e., $\mathcal{O}_{\{A_1,\ldots,A_n\}} = \{o \in \mathcal{O} \mid \forall i \in \{1,\ldots,n\} : A_i \notin del(o)\}$.

**Definition 19 (Ordering $\leq_E$ over Goal Sets)** *Let $(\mathcal{O}, I, \mathcal{G})$ be a planning problem with $\{A_1, \ldots, A_n\} \subseteq \mathcal{G}$ and $\{B_1, \ldots, B_k\} \subseteq \mathcal{G}$. Let $\mathsf{F}_{GP}^{\{A_1,\ldots,A_n\}}$ be the False set for $\{A_1, \ldots, A_n\}$. The ordering $\{B_1, \ldots, B_k\} \leq_E \{A_1, \ldots A_n\}$ holds if and only if*

$$\exists j \in \{1,\ldots,k\} : \forall o \in \mathcal{O}_{\{A_1,\ldots,A_n\}} : B_j \in add(o) \Rightarrow pre(o) \cap \mathsf{F}_{GP}^{\{A_1,\ldots,A_n\}} \neq \emptyset.$$

In a similar way, $\leq_h$ can be extended to $\leq_H$. For each $A_i$, the sets $\mathsf{F}_{DA}^{A_i}$ are determined based on Equation (3). The set $\mathsf{F}_{DA}^{\{A_1,\ldots,A_n\}}$ is simply the union over the individual sets:

$$\mathsf{F}_{DA}^{\{A_1,\ldots,A_n\}} := \bigcup_i \mathsf{F}_{DA}^{A_i} \qquad (4)$$

Then the fixpoint computation is entered with

$$\mathcal{O}^* := \mathcal{O} \setminus \{o \in \mathcal{O} \mid \exists\, i \in \{1,\ldots,n\} : A_i \in del(o) \vee \mathsf{F}_{DA}^{\{A_1,\ldots,A_n\}} \cap pre(o) \neq \emptyset\} \qquad (5)$$

The recomputation of $\mathcal{O}^*$ in each iteration of the fixpoint algorithm from Figure 1 is done accordingly. Apart from this, the algorithm remains unchanged.

**Definition 20 (Ordering $\leq_H$)** *Let $(\mathcal{O}, I, \mathcal{G})$ be a planning problem with $\{A_1,\ldots,A_n\} \subseteq \mathcal{G}$ and $\{B_1,\ldots,B_k\} \subseteq \mathcal{G}$. Let $\mathcal{O}^*$ be the set of actions that is obtained by performing the fixpoint computation shown in Figure 1, modified to handle sets of facts as defined in Equations (4) and (5). The ordering $\{B_1,\ldots,B_k\} \leq_H \{A_1,\ldots,A_n\}$ holds if and only if*

$$\exists\, j \in \{1,\ldots,k\} : \neg p\mathcal{A}(B_j, \mathcal{O}^*)$$

All given goal sets then undergo goal analysis, i.e., each pair of sets is checked for an ordering relation $\leq_E$ or $\leq_H$. Each derived relation defines an edge in a graph with the subgoal sets as nodes. The transitive closure is determined as before, and the degree of each node is computed. If the graph contains no disconnected nodes, then a total ordering over subsets of goals results by ordering the nodes based on their degree. This ordering defines the goal agenda. In the case of disconnected nodes, we default to the heuristic of adding the corresponding goals to the last goal set in the agenda.

### 4.3 The Agenda-Driven Planning Algorithm

Given a planning problem $(\mathcal{O}, \mathcal{I}, \mathcal{G})$, let us assume that a goal agenda $G_1, G_2, \ldots, G_k$ with $k$ entries has been returned by the analysis. Each entry contains a subset $G_i \subseteq \mathcal{G}$. The basic idea for the agenda-driven planning algorithm is now to first feed the planner with the original initial state $\mathcal{I}_1 := \mathcal{I}$ and the goals $\mathcal{G}_1 := G_1$, then execute the solution plan $\mathcal{P}$ in $\mathcal{I}$, yielding the new initial state $\mathcal{I}_2 = Result(\mathcal{I}_1, \mathcal{P})$. Then, a new planning problem is initialized as $(\mathcal{O}, \mathcal{I}_2, \mathcal{G}_2)$. After solving this problem, we want the goals in $G_2$ to be TRUE, but we also want the goals in $\mathcal{G}_1$ to remain TRUE, so we set $\mathcal{G}_2 := \mathcal{G}_1 \cup G_2$. The continuous merging of successive entries from the agenda yields a sequence of incrementally growing goal sets for the planner, namely

$$\mathcal{G}_i := \bigcup_{j=1}^{i} G_j$$

In a little more detail, the agenda-driven planning algorithm we implemented for IPP works as follows. First, IPP is called on the problem $(\mathcal{O}, \mathcal{I}, \mathcal{G}_1)$ and returns the plan $\mathcal{P}_1$, which achieves the subgoal set $\mathcal{G}_1$. $\mathcal{P}_1$ is a sequence of parallel sets of actions, which is returned by IPP similarly to GRAPHPLAN. Given this plan, the resulting state $R(\mathcal{I}, \mathcal{P}_1) = \mathcal{I}_2$ is computed based on the operational semantics of the planning actions.[6] In the case of a set of STRIPS actions, one simply adds all ADD effects to and deletes all DEL effects from a state description in order to obtain the resulting state, following the *Result* function in Definition 2. For STRIPS, the *Result* function coincides directly with the $R$ function. In the case of a set of parallel ADL actions, one needs to consider all possible linearizations of the parallel action set and has to deal with the conditional effects separately. For each linearization, a different resulting state can be obtained, but each of them will satisfy the goals. To obtain the new initial state $\mathcal{I}_2$, one takes the intersection of the resulting states for each possible linearization of the actions in a parallel set. This means to compute $n!$ linearizations for a parallel action set of $n$ actions in each time step. Since $n$ is usually small (more than 5 or 6 ADL actions per time step are very rare), the practical costs for this computation are neglectible.

This way, given a solution to a subproblem $(\mathcal{O}, \mathcal{I}_i, \mathcal{G}_i)$, one calculates the new initial state $\mathcal{I}_{i+1}$ and runs the planner on the subsequent planning problem $(\mathcal{O}, \mathcal{I}_{i+1}, \mathcal{G}_{i+1})$ until the planning problem $(\mathcal{O}, \mathcal{I}_k, \mathcal{G}_k)$ is solved.

The plan solving the original planning problem $(\mathcal{O}, \mathcal{I}, \mathcal{G})$ is obtained by taking the sequence of subplans $\mathcal{P}_1, \mathcal{P}_2, \ldots, \mathcal{P}_k$. One could argue that planning for increasing goal sets can lead to highly non-optimal plans. But IPP still uses the "no-ops first" strategy to achieve goals, which was originally introduced in the GRAPHPLAN system (Blum & Furst, 1997). Employing this strategy, the GRAPHPLAN algorithm, in short, first tries to achieve goals by simply keeping them TRUE, if possible. Since all goals $\mathcal{G}_1, \mathcal{G}_2, \ldots, \mathcal{G}_i$ are already satisfied in the initial state $\mathcal{I}_{i+1}$, starting from which the planner tries to achieve $\mathcal{G}_{i+1}$, this strategy ensures that these goals are only destroyed and re-established if no solution can be found otherwise. The no-ops first strategy is merely a GRAPHPLAN feature, but any reasonable planning strategy should preserve goals that are already true in the initial state whenever possible.

The soundness of the agenda-driven planning algorithm is obvious because $\mathcal{G}_k = \mathcal{G}$ and we have a sequence of sound subplans yielding a state transition from the initial state $\mathcal{I}$ to a state satisfying $\mathcal{G}$.

The completeness of the approach is less obvious and holds only if the planner cannot make wrong decisions before finally reaching the goals. More precisely, the approach is complete on problems that do not contain *deadlocks* as they were introduced in Definition 14.

**Theorem 8** *Given a solvable planning problem $(\mathcal{O}, I, \mathcal{G})$, and a goal agenda $\mathcal{G}_1, \mathcal{G}_2, \ldots \mathcal{G}_k$ with $\mathcal{G}_i \subseteq \mathcal{G}_{i+1}$ and $\mathcal{G}_k = \mathcal{G}$. Running any complete planner in the agenda-driven manner described above will yield a solution if the problem is deadlock-free.*

**Proof:** Let us assume the planner does not find a solution in step $i$ of the agenda-driven algorithm, i.e., no solution is found for the subproblem $(\mathcal{O}, \mathcal{I}_i, \mathcal{G}_i)$. As the planner is assumed to be complete on each subproblem, this implies unsolvability of $(\mathcal{O}, \mathcal{I}_i, \mathcal{G}_i)$. If this problem is not solvable, then neither is the problem $(\mathcal{O}, \mathcal{I}_i, \mathcal{G})$ solvable, since $\mathcal{G}_i \subseteq \mathcal{G}$ holds. Therefore, the goals cannot be reached from $\mathcal{I}_i$. Furthermore, $\mathcal{I}_i$ is a reachable state—it was reached by executing the partial solution plans $\mathcal{P}_1, \ldots, \mathcal{P}_{i-1}$ in the initial state. Consequently, $\mathcal{I}_i$ must be a deadlock state in the sense of Definition 14, which is a contradiction. ∎

---

6. See (Koehler et al., 1997) for the exact definition of $R$, which we do not want to repeat here.

This result states the feasibility of our approach: As we have shown, most benchmark problems that are currently investigated do contain inverse actions, are therefore invertible (Theorem 7), and are with that also deadlock-free (Theorem 6). Thus, with Theorem 8, our approach preserves completeness in these domains.

However in the general case, completeness cannot be guaranteed. The following example illustrates a situation where the assumption $s_{(A, \neg B)} \not\models p$ (assuming that preconditions of achieving actions are not contained in the state where $A$ is reached, cf. the derivation of the ordering $\leq_h$ in Section 3) is wrong and yields a goal ordering under which no plan can be found anymore although the problem is solvable.

Given the initial state $\{C, D\}$ and the goals $\{A, B\}$, the planner has the following set of ground STRIPS actions :

**op1:** $\{C\} \longrightarrow$ ADD $\{B\}$ DEL $\{D\}$
**op2:** $\{D\} \longrightarrow$ ADD $\{E\}$
**op3:** $\{E\} \longrightarrow$ ADD $\{F\}$
**op4:** $\{F\} \longrightarrow$ ADD $\{A\}$

The analysis will return an ordering $B \leq_h A$ because $B$ is only added by **op1**, but its precondition $C$ is not an effect of any of the other actions. Thus it concludes that $C$ is not reachable from a state in which $A$ holds. But in this example, $C$ holds in all reachable states. The assumption $s_{(A, \neg B)} \not\models C$ as made by the test $p\mathcal{A}(B, \mathcal{O}^*)$ is wrong. Thus, $B$ can be reached after $A$. On the other hand, $A \leq_r B$ holds, we even have a forced ordering $A \leq_f B$. But when testing for $A \leq_h B$, this ordering remains undetected, because our method does not discover that the precondition $F$ of **op4** is not achievable from the state in which $B$ holds: we obtain $\mathsf{F}_{DA}^B = \{D\}$, which excludes **op2** from $\mathcal{O}^*$, but **op3** and **op4** remain in the set of usable actions. Thus, **op4** is considered a legal achiever of $A$, and **op3** is considered a legal achiever for its precondition $F$. We could only detect the right ordering if we regressed over the action chain **op4**, **op3**, **op2** and found out that, with $D$ being in the F set of $B$, all these actions must be excluded from $\mathcal{O}^*$.

Consequently, the goal agenda $\{B\}, \{A\}$ is fed into the planner, which solves the first subproblem using **op1**, but then fails in achieving $A$ from the state $\{B, C\}$ since there is no inverse action to **op1** and $D$ cannot be re-established in any other way.

## 5. Empirical Results

We implemented both methods to approximate $\leq_r$ as a so-called Goal Agenda Manager (GAM) for the IPP planning system (Koehler et al., 1997). GAM is activated after the set of ground actions has been determined and either uses $\leq_e$ or $\leq_h$ to approximate the reasonable goal ordering. Then it calls the IPP planning algorithm on each entry from the goal agenda and outputs the solution plan as the concatenation of the solution plans that have been found for each entry in the agenda.[7]

---

7. The source code of GAM, which is based on IPP 3.3, and the collection of domains from which we draw the subsequent examples can be downloaded from http://www.informatik.uni-freiburg.de/~koehler/ipp/gam.html. All experiments have been performed on a SPARC 1/170.

The empirical evaluation that we performed uses the IPP domain collection, which contains 48 domains with more than 500 planning problems. Out of these domains, we were able to derive goal ordering information in 10 domains. These domains indeed pose constraints on the ordering in which a planner has to a achieve a set of goals. In all other domains, where no goal orderings could be derived, we found that either only a single goal has to be achieved, for example in the *manhattan*, *movie*, *molgen*, and *montlake* domains or the goals can be achieved in any order, as for example in the *logistics*, *gripper*, and *ferry* domains. We found no benchmark domain, in which a natural goal ordering existed, but our method failed to detect it. As a matter of fact, looking at a goal ordering that seems to be natural, one usually finds that the ordering is reasonable in the sense of Definition 8, see for example the *blocks world*, *woodshop*, and *tyreworld* domains. Our method finds almost all of the reasonable orderings, which indicates that both approximation techniques $\leq_e$ and $\leq_h$ are appropriate for detecting ordering information.

In the following, we will first compare the $\leq_e$ and $\leq_h$ techniques in terms of runtime and number of goal agenda entries generated. Then we take a closer look at the agendas that are generated in selected domains and investigate how they influence the performance of the IPP planning system. The exact definition of all domains can be downloaded from the IPP webpage, we just give the name of the domain and the name of the particular planning problem as well as the number of (ground) actions a domain contains, because this parameter nicely characterizes the size of a domain and with that usually the difficulty to handle it.

In all examples, the times shown to compute the goal agenda contain the effort to parse and instantiate the operators, i.e., to compute the set of actions. Times for parsing and instantiation are not listed explicitly, because they are, on the test examples used here, usually very close to zero and do not influence the performance of the planner in a significant way.

### 5.1 Comparison of $\leq_h$ and $\leq_e$

We begin our comparison with a summary of results that we obtained in different representational variants of the *blocks world*. The *bw_large_a* to *bw_large_d* examples originate from the SATPLAN test suite (Kautz & Selman, 1996) to which we added the larger examples *bw_large_e* to *bw_large_g*. The *parcplan* example comes from (El-Kholy & Richards, 1996) and uses multiple grippers and limited space on the table. The *stack_n* examples use the GRAPHPLAN *blocks world* representation and simply require to stack $n$ blocks on each other, which are all on the table in the initial state.

The two methods return exactly the same ordering relations across all *blocks world* problems. But as Figure 5 confirms, the computation of $\leq_e$ based on planning graphs is much more time-consuming. It hits the computational border when a domain contains more than 10000 actions. The computation of $\leq_h$ is much faster and also scales to larger action sets.

| problem | #actions | #agenda entries | CPU($\leq_e$) | CPU($\leq_h$) |
|---|---|---|---|---|
| bw_large_a | 162 | 1 | 0.69 | 0.07 |
| bw_large_b | 242 | 5 | 1.45 | 0.11 |
| bw_large_c | 450 | 7 | 4.85 | 0.22 |
| bw_large_d | 722 | 11 | 14.18 | 0.35 |
| bw_large_e | 722 | 11 | 12.95 | 0.35 |
| bw_large_f | 1250 | 6 | 44.93 | 0.58 |
| bw_large_g | 1800 | 9 | 97.11 | 0.88 |
| parcplan | 1960 | 4 | 25.84 | 1.47 |
| stack_20 | 800 | 19 | 6.91 | 0.36 |
| stack_40 | 3200 | 39 | 160.00 | 1.74 |
| stack_60 | 7200 | 59 | 840.42 | 4.85 |
| stack_80 | 12800 | 79 | - | 11.38 |

Figure 5: Comparison of $\leq_e$ and $\leq_h$ on *blocks world* problems. #actions shows the number of actions in the set $\mathcal{O}$, from which the planner tries to construct a plan. #agenda entries says how many goal subsets have been detected and ordered by GAM. Column 4 and 5 display the CPU time that is required by both methods to compute the agenda when provided with the set $\mathcal{O}$. A dash will always mean that IPP ran out of memory on a 1 Gbyte machine.

Figure 6 and Figure 7 show the results for the other domains, in which our method is able to detect reasonable orderings. Figure 6 lists the domains, in which both methods return the same goal agendas. The *tyreworld*, *hanoi*, and *fridgeworld* domains originate from UCPOP (Penberthy & Weld, 1992), while the *link-repeat* domain can be found in (Veloso & Blythe, 1994). The performance results coincide with those shown in Figure 5. Figure 7 shows the same picture in terms of runtime performance, but in these domains different agendas are returned by $\leq_e$ and $\leq_h$.

The *woodshop* and *scheduling* domains contain actions with conditional effects, while the other domains only use STRIPS operators. The computation of $\leq_e$ fails to derive goal orderings for all *scheduling world* problems (of which we only display the largest problem *sched6*) and for the *wood1* problem. The explanation for this behavior can be found in the different treatment of conditional effects by both methods. IPP does only find a very limited form of *mutex* relations between conditional effects when building the planning graph. A goal, which is achieved with a conditional effect, will not very often be exclusive to a large number of other facts in the graph. Thus, the F sets are very small or sometimes even empty and consequently, only very few actions can be excluded when performing the reachability analysis and thus, reasonable orderings may remain undetected. Direct analysis investigates the conditional effects in more detail and is therefore able to derive much larger F sets.

The behavior of the $\leq_h$ method in the STRIPS domains *bulldozer*, *glassworld*, and *shopping world* is caused by the same phenomenon. In these domains, one can derive much larger F sets using planning graphs and in turn these sets exclude more actions. Since direct analysis finds smaller or empty F sets, it also finds less $\leq_h$ relations. The *woodshop* domain

| domain | problem | #actions | #agenda entries | CPU($\leq_e$) | CPU($\leq_h$) |
|---|---|---|---|---|---|
| tyreworld | fixit1 | 26 | 6 | 0.05 | 0.01 |
|  | fixit2 | 59 | 6 | 0.20 | 0.03 |
|  | fixit3 | 108 | 6 | 0.45 | 0.06 |
|  | fixit4 | 173 | 6 | 0.84 | 0.10 |
|  | fixit5 | 254 | 6 | 1.56 | 0.15 |
|  | fixit10 | 899 | 6 | 16.29 | 0.64 |
| hanoi | hanoi3 | 48 | 3 | 0.05 | 0.02 |
|  | hanoi4 | 90 | 4 | 0.10 | 0.04 |
|  | hanoi5 | 150 | 5 | 0.19 | 0.08 |
|  | hanoi6 | 231 | 6 | 0.35 | 0.12 |
|  | hanoi7 | 336 | 7 | 0.63 | 0.19 |
| fridgeworld | fridge | 779 | 2 | 0.77 | 0.55 |
| link-repeat | link10 | 31 | 2 | 0.19 | 0.01 |
|  | link30 | 31 | 2 | 0.21 | 0.01 |

Figure 6: Comparison of $\leq_e$ and $\leq_h$ on those benchmark domains, in which they return identical agendas.

| domain | problem | #actions | #agenda entries | CPU($\leq_e$) | CPU($\leq_h$) |
|---|---|---|---|---|---|
| bulldozer | bull | 61 | 2/1 | 0.09 | 0.03 |
| glassworld | glass1 | 26 | 2/1 | 0.02 | 0.01 |
|  | glass2 | 114 | 2/1 | 0.19 | 0.09 |
|  | glass3 | 122 | 2/1 | 0.22 | 0.09 |
| shoppingworld | shop | 81 | 2/1 | 0.07 | 0.02 |
| scheduling | sched6 | 104 | 1/4 | 01.0 | 0.12 |
| woodshop | wood1 | 15 | 1/3 | 0.03 | 0.01 |
|  | wood2 | 15 | 6/5 | 0.03 | 0.01 |
|  | wood3 | 43 | 6/5 | 0.14 | 0.06 |

Figure 7: Domains in which $\leq_e$ and $\leq_h$ return different goal agendas, which we give in the form $n_1/n_2$. The number before the slash says how many entries are contained in the agenda computed by $\leq_e$, the number following the slash says how many entries are contained in the agenda computed by $\leq_h$. #agenda entries=1 means that the agenda contains only a single entry, namely the original goal set, and no ordering was derived.

shows that the results can differ within the same domain, but depending on the specific planning problem. The problem *wood2* varies from the problem *wood1* in the sense that one goal is slightly different—an object needs to be put into a different shape—and that two more goals are present. While there are no goal orderings derived between pairs of the old

371

goals from *wood1*, lots of $\leq_e$ relations are derived between mixed pairs of old and new goals in *wood2*, yielding a detailed goal agenda. The problem *wood3* contains additional objects and many more goals, which can also be successfully ordered.

In the subsequent experiments, we decided to solely use the heuristic ordering $\leq_h$ because the computation of $\leq_h$ is less costly than the computation of $\leq_e$ in all cases, yielding comparable agendas in most cases. In the three domains that we investigate more closely, namely the *blocks world*, *tyreworld* and *hanoi* domains, the agendas derived by both methods are, in fact, exactly the same.

### 5.2 Influence of Goal Orderings on the Performance of IPP and Interaction with RIFO

In this section, we analyze the influence of the goal agenda on the performance of IPP and combine it with another domain analysis method, called RIFO (Nebel, Dimopoulos, & Koehler, 1997). RIFO is a family of heuristics that enables IPP to exclude irrelevant actions and initial facts from a planning problem. It can be very effectively combined with GAM, because if IPP plans for only a subset of goals from the original goal set, it is very likely that also only a subset of the relevant actions is needed to find a plan. More precisely, we obtain one subproblem for each entry in the agenda, and, for each such subproblem, we use RIFO for preprocessing before planning with IPP. In this configuration, GAM reduces the search space for IPP by decreasing the number of subgoals the planner has to achieve at each moment, while RIFO reduces the search space dramatically by selecting only those actions that are relevant for this goal subset.

#### 5.2.1 The Blocks World

Figure 8 illustrates the *parcplan* problem (El-Kholy & Richards, 1996) in detail. Seven robot arms can be used to order 10 blocks into 3 stacks on 5 possible positions on the table.

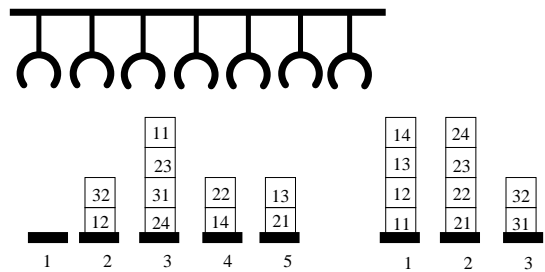

Figure 8: The *parcplan* problem with limited space on the table, seven robot arms, and several stacks.

The goal agenda derived by IPP orders the blocks into horizontal layers:

1: *on-table(21, t2)* ∧ *on-table(11, t1)*
2: *on-table(31, t3)* ∧ *on(22, 21)* ∧ *on(12, 11)*
3: *on(32, 31)* ∧ *on(13, 12)* ∧ *on(23, 22)*
4: *on(14, 13)* ∧ *on(24, 23)*

The optimal plan of 20 actions solving the problem is found by IPP using GAM in 14 s, where it spends one second on computing the goal agenda, almost 13 seconds to build the planning graphs, but only 0.01 second to search for a plan. Only 70 actions have to be tried to find the solution. Without the goal analysis, IPP needs approx. 47 s and searches 52893 actions in more than 26 seconds.

RIFO (Nebel et al., 1997) fails in detecting a subset of relevant actions when the original goal set has to be considered, but it succeeds in selecting relevant actions for the subproblems stated in the agenda. It reduces runtime down to less than 8 s with 1 s again spent on the goal agenda, almost 6 s spent on the removal of irrelevant actions and initial facts, less than 1 s spent on building the planning graphs. As previously, almost no time is spent on planning.

Figure 9 shows IPP on the SATPLAN *blocks world* examples from (Kautz & Selman, 1996), the *bw_large.e* example taken from (Dimopoulos, Nebel, & Koehler, 1997), and two very large examples *bw_large.f* (containing 25 blocks and requiring to build 6 stacks in the goal state) and *bw_large.g* with 30 blocks/8 stacks.

| SATPLAN | # actions | plan length | IPP | +G | +G+R | +G+R+L |
|---|---|---|---|---|---|---|
| bw_large.a | 162 | 12 (12) | 0.70 | 0.74 | 0.58 | 0.34 |
| bw_large.b | 242 | 22 (18) | 26.71 | 0.86 | 0.55 | 0.52 |
| bw_large.c | 450 | 48 | - | 7.34 | 2.42 | 2.58 |
| bw_large.d | 722 | 54 | - | 11.62 | 3.74 | 3.81 |
| bw_large.e | 722 | 52 | - | 11.14 | 3.99 | 3.97 |
| bw_large.f | 1250 | 90 | - | - | - | 16.01 |
| bw_large.g | 1800 | 84 | - | - | 117.56 | 28.71 |

Figure 9: Performance on the extended SATPLAN *blocks world* test suite. The second column shows the number of ground actions in this domain, the third column shows the plan length, i.e., the number of actions contained in the plan, generated by GAM and in parentheses the plan length generated by IPP without GAM given that IPP without GAM is able to solve the corresponding problem. +G means that IPP is using GAM, +G+R means IPP uses GAM and RIFO, +G+R+L means that subgoals from the same set in the agenda are arbitrarily linearized. All runtimes cover the whole planning process starting with parsing the operator and domain file, performing the GAM and RIFO analysis (if active), and then searching the graph until a plan is found.

IPP 3.3 without GAM can only solve the *bw_large.a* and *bw_large.b* problems. Using a goal agenda, some plans become slightly longer, but performance is increasing dramatically. Plan length is growing because blocks are accidentally put in positions where they cut off goals that are still ahead in the agenda and thus, additional actions need to be added to the plan to remove these blocks from wrong positions. A further speed-up is possible when RIFO is additionally used, because it reduces the size of planning graphs dramatically. Finally, goals that belong to the same subset in the agenda can be linearized based on the heuristic assumption that if the analysis found no reasonable goal orderings, then the goals are achievable in any order. With this option, the problems are solved almost instantly.

The reader may wonder at this point why we use linearization of agenda entries only as an extra option and do not investigate it further. There are two reasons for that. First, linearization does have negative side effects in most domains that we investigated. For example, it yields much longer plans in the *logistics* domain and all its variants. When linearizing the single entry that the agenda for a *logistics* problem contains, all packages get transported to their goal position one by one. Of course, this takes much more planning steps than simultaneously transporting packages with coinciding destinations.

Secondly, the effects of linearization are somewhat unpredictable, even in domains where it usually tends to yield good results. This is because GAM does not recognise *all* interactions between goals. Consider a *blocks world* problem with four blocks $A$, $B$, $C$ and $D$. Say $B$ is positioned on $C$ initially, the other blocks being each on the table, and the goal is to have $on(A, B)$ and $on(C, D)$. The agenda for this problem will comprise a single entry containing both goals. In fact, there is no reasonable goal ordering here. Nevertheless, stacking $A$ onto $B$ immedeatly is a bad idea, as the planner needs to move $C$ to achieve $on(C, D)$. Being not aware of this, GAM might linearize the single agenda entry to have $on(A, B)$ up front, which makes the problem harder than it actually is. Thus, the runtime advantages that linearization sometimes yields on the *blocks world* can be more or less seen as cases of "good luck".

Figure 10 shows IPP on the *stack_n* problems. IPP without any domain analysis can handle up to 12 blocks in less than 5 minutes, but for 13 blocks more than 15 minutes are needed. Using GAM, 40 blocks can be stacked in less than 5 minutes. Using GAM and RIFO, the 5 minutes limit is extended to 80 blocks, while *stack100* is solved in 11.5 min where 11.3 min are spent for both analysis methods and only 0.2 min are needed for building the planning graphs and extracting a plan.

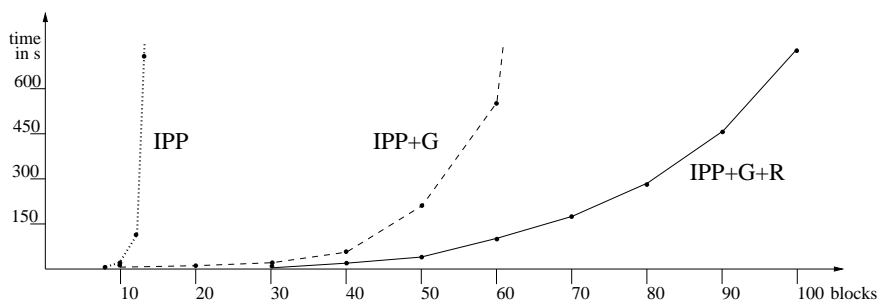

Figure 10: IPP 3.3 on a simple, but huge stacking problem.

Figure 11 shows the sharing of the overall problem-solving time between GAM, RIFO and the IPP search algorithm on *blocks world* problems. Similar results are obtained in the *tyreworld*. GAM takes between 3 and 16 %, RIFO takes between 75 and 96 %, and the search effort is reduced down to approx. 1 %. The overall problem solving time is clearly determined by RIFO, while the search effort becomes a marginal factor in the determination of performance. This indicates that a further speed-up is possible when improving the performance of GAM and RIFO. It also indicates that even the hardest planning problems can become easy if they are structured and decomposed in the right way.

| problem  | # actions | GAM          | RIFO           | search algorithm |
|----------|-----------|--------------|----------------|------------------|
| stack_20 | 800       | 0.31 = 16 %  | 1.44 = 75 %    | 0.13 = 7 %       |
| stack_40 | 3200      | 1.57 = 7 %   | 18.77 = 90 %   | 0.51 = 2 %       |
| stack_60 | 7200      | 4.40 = 4 %   | 93.10 = 94 %   | 1.15 = 1 %       |
| stack_80 | 12800     | 9.60 = 3 %   | 283.60 = 96 %  | 2.33 = 1 %       |
| parcplan | 1960      | 0.86 = 12 %  | 5.52 = 76 %    | 0.83 = 11 %      |

Figure 11: Distribution of problem-solving time on *blocks world* examples between GAM, RIFO, and the search algorithm, which comprises the time to build and search the planning graph. The remaining fraction of total problem-solving time, which is not shown in the table, is spent on parsing and instantiating the operators.

5.2.2 THE TYREWORLD

The tyreworld problem, originally formulated by Stuart Russell, asks a planner to find out how to replace a flat tire. It is easily solved by IPP within a few milliseconds. The problem becomes much harder if the number of flat tires is increasing, cf. Figure 12.

| Tires | # actions | IPP          | +G+R         | +G+R+L         | Search Space |
|-------|-----------|--------------|--------------|----------------|--------------|
| 1     | 26        | 0.10 (12/19) | 0.15 (14/19) | 0.16 (17/19)   | 1298/88      |
| 2     | 59        | 17.47 (18/30)| 0.41 (24/32) | 0.32 (30/34)   | 1290182/210  |
| 3     | 108       | -            | 2.87 (32/44) | 0.63 (41/46)   | -/366        |
| 4     | 173       | -            | -            | 1.12 (52/60)   | -/565        |
| 5     | 254       | -            | -            | 1.93 (63/73)   | -/807        |
| 6     | 353       | -            | -            | 3.42 (73/85)   | -/1092       |
| 7     | 464       | -            | -            | 4.81 (84/98)   | -/1420       |
| 8     | 593       | -            | -            | 8.07 (95/121)  | -/1791       |
| 9     | 738       | -            | -            | 11.27 (106/124)| -/ 2205      |
| 10    | 899       | -            | -            | 16.89 (118/136)| -/2662       |

Figure 12: IPP in the Tyreworld. The numbers in parentheses show the time steps, followed by the number of actions in the generated plan. The last column compares the search spaces. The number before the slash shows the "number of actions tried" parameter for the plain IPP planning algorithm, while the number following the slash shows the "number of actions tried" for IPP using GAM, RIFO, and the linearization of entries in the agenda. A dash means that the "number of actions tried" is unknown because IPP failed in solving the corresponding planning problem.

IPP is only able to solve the problem for 1 and 2 tires. Using GAM and RIFO, 3 tires can be handled. Solution length under GAM is slightly increasing, which is caused by superfluous *jack-up* and *jack-down* actions. In short, this is explained as follows. Each wheel needs to be mounted on its hub, which is expressed by an *on(?r, ?h)* goal. To mount a wheel, its hub must be jacked up. After mounting, the nuts are done up. Then, the hub needs to be jacked down again, in order to tighten the nuts achieving a *tight(?n, ?h)* goal. Now, GAM puts all of the *on* goals into one entry preceeding the *tight* goals. Thus, solving the entry containing the *on* goals, each hub is jacked up, the wheel is put on, and the hub is immediatly jacked down again in order to replace the next wheel. Afterwards, solving the *tight* goals, each hub must be jacked up—and down—one more time for doing up the nuts. Solving the problem in this manner, the planner inserts one superfluous *jack-up*, and one superfluous *jack-down* action for each wheel. More precisely, superfluous actions are inserted for all but one wheel, namely the wheel that is last mounted when solving the *on* goals. After mounting this wheel, all *on* goals are achieved, and the planner proceeds to the next agenda entry with this wheel still being jacked up. Then, trying to achieve the *tight* goals, IPP recognizes that the shortest plan (in terms of the number of parallel steps) results when the nuts are first done up on the hub that is already jacked up. Thus, this hub is only jacked up *one* time, achieving the corresponding *on* goal, and jacked down again one time, before achieving its *tight* goal.

In the case of 3 tires, the following goal subsets are identified and ordered:

1: inflated(r3), inflated(r2), inflated(r1)
2: on(r3, hub3), on(r1, hub1), on(r2, hub2)
3: tight(n2, hub2), tight(n3, hub3), tight(n1, hub1)
4: in(w3, boot), in(pump, boot), in(w1, boot), in(w2, boot)
5: in(jack, boot)
6: in(wrench, boot)
7: closed(boot)

The hardest subproblem in the agenda is to achieve the $on(r_i, hub_i)$ goals in entry 2, i.e., to mount inflated spare wheels on the various hubs. Trying to generate a maximum parallelized plan is impossible for IPP for more than 3 tires. But since the goals are completely independent of each other, any linearization of them will perfectly work. The resulting plans become slightly longer due to the way that the *tight* goals are achieved when using the -L option. We noticed earlier that for one wheel (the one that is last mounted when solving the *on* goals) no superfluous *jack-up* and *jack-down* actions need to be inserted into the plan. Linearizing the agenda entries, superfluous *jack-up* and *jack-down* actions must most likely be inserted for *all* wheels, yielding plans that are two steps longer. The reason for that is that any *tight* goal might be the first in the linearization. Most likely, this is not the *tight* goal corresponding to the hub that is still jacked up, so the planner needs to insert one superfluous *jack-down* action here. Later, it must jack up this hub again, yielding another superfluous action. Using +G+R+L in the case of 10 tires, only 2662 actions need to be tried until a plan of 136 actions is found, which takes 0.08 s. GAM requires 0.55 s, RIFO requires 14.42 s, 1.74 s are consumed to generate the planning graphs, and 0.08 s are spent to compute the initial states for all subproblems. The remaining 0.02 s are consumed for parsing and instantiating.

### 5.2.3 THE TOWER OF HANOI

A surprising result is obtained in the *tower of hanoi* domain. In this domain, a stack of discs has to be moved from one peg to a third peg with an auxiliary second peg between them, but never a larger disc can be put onto a smaller disc. In the case of three discs $d1$, $d2$, $d3$ of increasing size, the goals are stated as *on(d3,peg3)*, *on(d2,d3)*, *on(d1,d2)*. GAM returns the following agenda, which correctly reflects the ordering that the largest disc needs to be put in its goal position first.

1: *on(d3,peg3)*

2: *on(d2,d3)*

3: *on(d1,d2)*

The goal agenda leads to a partition into subproblems that corresponds to the recursive formulation of the problem solving algorithm, i.e., to solve the problem for $n$ discs, the planner first has to solve the problem for $n-1$ discs, etc. For the first entry, a plan of 4 actions (time steps 0 to 3 below) is generated, which achieves the goal *on(d3,peg3)*.[8] Then a plan of 2 actions (time steps 4 and 5) achieves the goals *on(d3,peg3)* and *on(d2,d3)* with *on(d3,peg3)* holding already in the initial state. Finally, a one-step plan (time step 6) is generated that moves the third disc with the other two discs being already in the goal position.

```
time step 0:   move(d1,d2,peg3)
time step 1:   move(d2,d3,peg2)
time step 2:   move(d1,peg3,d2)
time step 3:   move(d3,peg1,peg3)

time step 4:   move(d1,d2,peg1)
time step 5:   move(d2,peg2,d3)
time step 6:   move(d1,peg1,d2)
```

Surprisingly, IPP is not able to benefit from this information, but runtime of IPP using GAM is exploding dramatically for increasing numbers of discs, see Figure 13.

| discs | #actions | IPP | IPP +G | UCPOP | UCPOP on subproblems |
|---|---|---|---|---|---|
| 2 | 21 | 0.02 | 0.02 | 0.12 (27) | 0.06 (17) + 0.02 (6) |
| 3 | 48 | 0.08 | 0.07 | 8.00 (2291) | 0.18 (48) + 0.06 (13) + 0.01 (6) |
| 4 | 90 | 0.33 | 0.25 | - | - |
| 5 | 150 | 1.57 | 3.10 | - | - |
| 6 | 231 | 9.71 | 88.45 | - | - |
| 7 | 336 | 69.44 | 2339.94 | - | - |

Figure 13: Runtimes of IPP with and without the goal agenda on *hanoi* problems compared to UCPOP without agenda and UCPOP on the agenda subproblems using ZLIFO and the *ibf* control strategy.

---

8. A *move* action takes as first argument the disc to be moved, as second the disc from which it is moved, and as third argument the disc or peg to which it is moved.

We are not able to provide an explanation for this phenomenon, but the division into subproblems causes a much larger search space for the planner although the same solution plans result. RIFO cannot improve on the situation because it selects all actions as relevant.

The *tower of hanoi* domain is the only one we found where IPP's performance is deteriorated by GAM. We do currently not see a way of how one can tell in advance whether IPP will gain an advantage from using GAM or not. The overhead caused by the goal analysis itself is very small, but an "inadequate" split of the goals into subgoal sets can lead to more search, see also Section 6.

However in this case, the phenomenon seems to be specific to IPP. We simulated the information that is provided by GAM in UCPOP and obtained a quite different picture. The fifth column in Figure 13 shows the runtime of UCPOP using ZLIFO (Pollack, Joslin, & Paolucci, 1997) and the *ibf* control strategy with the number of explored partial plans in parentheses. UCPOP can only solve the problem for 2 and 3 discs. In the last column of the figure, we show the runtime and number of explored partial plans, which result when UCPOP is run on the subproblems that result from the agenda. These are exactly the same subproblems which IPP has to solve, but the performance of UCPOP improves significantly. Instead of taking 8 s and exploring 2291 partial plans, UCPOP only takes 0.18+0.06+0.01=0.25 s and explores only 48+13+6=67 plans. Unfortunately, any problems or subproblems with more than 3 discs remain beyond the performance of UCPOP. The performance improvement is independent of the search strategies used by UCPOP. For example, if *ibf* control is used without ZLIFO, the number of explored partial plans is reduced from 78606 down to 2209 in the case of the problem with 3 discs. Runtime improves from 65 seconds down to 2 seconds. Similarly, when using *bf* control without ZLIFO the number of explored partial plans reduces from 1554 down to 873.

Knoblock (1994) also reports an improvement in performance for the Prodigy planner (Fink & Veloso, 1994) when it is using the abstraction hierarchy generated for this domain by the ALPINE module, which provides in essence the same information as the goal agenda.[9]

## 6. Summary and Comparison to Related Work

Many related approaches have been developed to provide a planner with the ability to decompose a planning problem by giving it any kind of goal ordering information. Subsequently, we discuss the most important of them and review our own work in the light of these approaches.

Our method introduces a preprocessing approach, which derives a total ordering for subsets of goals by performing a static, heuristic analysis of the planning problem at hand. The approach works for domains described with STRIPS or ADL operators and is based on polynomial-time algorithms. The purpose of this method is to provide a planner with search control, i.e., we opt at deriving a goal achievement order and then successively call the planner on the totally ordered subsets of goals.

The method preserves the soundness of the planning system, but the completeness only in the case that the planning domain does not contain deadlocks. We argue that benchmark domains quite often possess this property, which is also supported by other authors (Williams & Nayak, 1997).

---

9. However, to find that goal ordering information, ALPINE requires to represent the tower of hanoi domain involving several operators, cf. (Knoblock, 1991).

The computation of $\leq_h$ and $\leq_e$ requires only polynomial time, but both methods are incomplete in the sense that they will not detect all reasonable goal orderings in the general case. The complexity of deciding on the existence of *forced* and *reasonable* goal orderings has been proven to be PSPACE-hard in Section 2 and therefore, trading completeness for efficiency seems to be an acceptable solution. Our complexity results relate to those found by Bylander (1992) who proves the PSPACE-completeness of serial decomposability (Korf, 1987). Given a set of subgoals, *serial decomposability* means that previously satisfied subgoals do not need to be violated later in the solution path, i.e., once a subgoal has been achieved, it remains valid until the goal is reached. The purpose of our method is to derive constraints that make those orderings explicit under which no serial decomposability of a set of goals can be found, i.e., we consider the complementary problem, which is also reflected in our complexity proofs.

In many cases, we found that the goal agenda manager can significantly improve the performance of the IPP planning system, but we found at least one domain, namely the *tower of hanoi*, where a dramatic decrease in performance can be observed although IPP still generates the optimal plan when processing the ordered goals from the agenda. So far, the complexity results of Bäckström and Jonsson (1995) predicted that planning with abstraction hierarchies can be exponentially less efficient, but because exponentially longer plans can be generated.

The idea to analyze the effects and preconditions of operators and to derive ordering constraints based on the interaction of operators can also be found in a variety of approaches. While we analyze harmful interactions of operators in our method by studying the delete effects, the approaches described in (Dawsson & Siklossy, 1977; Korf, 1985; Knoblock, 1994) concentrate on the positive interactions between operators. The successful matching of effects to preconditions forms the basis to learn *macro-operators*, see (Dawsson & Siklossy, 1977; Korf, 1985).

The ALPINE system (Knoblock, 1994) learns abstraction hierarchies for the Prodigy planner (Fink & Veloso, 1994). The approach is based on an ordering of the preconditions and the effects of each operator, i.e., all effects of an operator must be in the same abstraction hierarchy and its preconditions must be placed at the same or a lower level than its effects. This introduces an ordering between the possible subgoals in a domain, which is orthogonal to the ordering we compute: In ALPINE, a subgoal $A$ is ordered before a subgoal $B$ if $A$ *enables* $B$, i.e., $A$ must be *possibly* achieved first in order to achieve $B$. Our method orders $A$ before $B$ if $A$ cannot be achieved without *necessarily* destroying $B$. The result of ALPINE and GAM are a set of binary constraints. In the case of ALPINE, the constraints are computed between all atoms in a domain, while GAM restricts the analysis to the goals only. Both approaches represent the binary constraints in a graph structure. ALPINE merges atomic goals together if they belong to a strongly connected component in the graph. GAM merges sets of goals together if they have identical degree. Then they both compute a topological sorting of the sets that is consistent with the constraints. The resulting goal orderings can be quite similar as the examples by Knoblock (1994) demonstrate, but GAM approximates reasonable goal orderings in domains where ALPINE fails in finding abstraction hierarchies. Two further examples (Knoblock, 1991) are the *tower of hanoi* domain using only one *move* operator and the *blocks world*. In both domains, ALPINE cannot detect the orderings because it investigates the operator schemata, not the set of ground actions, and therefore cannot distinguish the orderings between different instantiations of the same literal. Although ALPINE could be modified to handle ground actions, this will significantly increase the amount of computation it requires. GAM on the other hand, handles large sets of ground actions in an efficient way, in particular if *direct analysis* is used.[10]

An analysis, which is quite similar to ALPINE, but which is performed in the framework of HTN planning, is described by Tsuneto et al. (1998). The approach analyzes the *external* conditions of methods, which cannot be achieved when decomposing the method further. This means, such conditions have to be established by the decomposition of those methods, which *precede* the method using this external condition. Two strategies to determine the decomposition order of methods are defined and empirically compared. Here lies the main difference to the other approaches described so far: Instead of trying to automatically construct the decomposition orderings, they are predefined and fixed for all domains and problems.

Harmful interactions among operators are studied by Smith and Peot (1993) and Etzioni (1993). A *threat* of an operator $o$ to a precondition $p$ occurs if there is an instantiation of $o$ such that its effects are inconsistent with $p$ (Smith & Peot, 1993). The knowledge about threats is used to control a plan-space planner. In contrast to a state-space planner such as IPP, computing an explicit ordering of goals does not prevent the presence of threats in a partial plan because the order in which the goals are processed does not determine the order in which actions occur in the plan. The notion of forced and reasonable goal orderings is not comparable to that of a threat because a threat still has the potential of being resolved by adding binding or ordering constraints to the plans. In contrast to this, a forced or reasonable goal ordering persists under all bindings and enforces a specific ordering of the subgoals.

Given a planning problem, STATIC (Etzioni, 1993) computes a backchaining tree from the goals in the form of an AND/OR graph, which it subsequently analyzes for the occurrence of goal interactions that will *necessarily* occur. This analysis is much more complicated than ours, because STATIC has to deal with uninstantiated operators and axioms, which describe properties of legal states. The result of the analysis are goal ordering rules, which order goals if certain conditions are satisfied in a state. This is the main difference to GAM, which generates explicit goal orderings independently of a specific state. It does not need to extract conditions that a specific state has to satisfy because it considers the generic state $s_{(A,\neg B)}$ in the analysis, which represents all states satisfying $A$, but not $B$. As GAM, STATIC is incomplete in the sense that it cannot detect all existing goal interactions. The problem for GAM is that deciding reasonable orderings is PSPACE-hard, as we have proven in this paper. The problem for STATIC is that it has to compute the *necessary* effects of an operator in a given state. As Etzioni (1993) conjectures and Nebel and Bäckström (1994) prove, this problem is computationally intractable and therefore, any polynomial-time analysis method must be incomplete.

---

10. Abstraction hierarchies are more general than the goal orderings we compute. They cannot only serve for the purpose of providing a planner with goal ordering information, but also allow to generate plans at different levels of refinement, see also (Bacchus & Yang, 1994). Two other approaches generating abstraction hierarchies based on numerical *criticality* values can be found in (Sacerdoti, 1974; Bundy, Giunchiglia, Sebastiani, & Walsh, 1996).

Last, but not least there have been quite a number of approaches in the late Eighties, which focused directly on subgoal orderings. These fall into two categories: The approaches described in (Drummond & Currie, 1989; Hertzberg & Horz, 1989) focus on the detection of conflicts caused by goal interdependencies to guide a partial-order planner during search. We do not investigate these approaches in more detail here because they do not extract explicit goal orderings as a preprocess to planning as we do. The works described in (Irani & Cheng, 1987; Cheng & Irani, 1989; Joslin & Roach, 1990) implement preprocessing approaches, which perform a structural analysis of the planning task to determine an appropriate goal ordering before planning starts. Irani and Cheng (1987) compute a relation $\prec$ between pairs of goals, which—roughly speaking—orders a goal $A$ after a goal $B$ if $B$ must be achieved before $A$ can be achieved. Their formalism is rather complicated and the theoretical properties of the relation are not investigated. In (Cheng & Irani, 1989), the approach is extended such that *sets* of goals can be ordered with respect to each other. The exact properties of the formalism remain unclear. In (Joslin & Roach, 1990), a graph-theoretical approach is described that generates a graph with all atoms from a given domain description as nodes and draws an arc between a node $A$ and a node $B$ if an operator exists that takes $A$ as precondition and has $B$ as an effect. When assuming that all operators have inverse counterparts, identifying connected components in the graph is proposed as a means to order goals. The approach is unlikely to scale to the size of problem spaces today's planners consider and it is also completely outdated in terms of terminology.

Finally, one can wonder how the reasonable and forced goal orderings relate to others defined in the literature. There is only one attempt of which we know where an ordering relation is explicitly defined and its properties are studied, see (Hüllem et al., 1999). In this paper, the notion of *necessary* goal orderings is introduced, which must be true in all *minimal solution plans* (Kambhampati, 1995).[11] The approach extends operator graphs (Smith & Peot, 1993) and orders a goal based on three criteria called *goal subsumption*, *goal clobbering*, and *precondition violation*. Goal subsumption $A < B$ holds if every solution plan achieving a goal $B$ in a state $s$ also achieves a goal $A$ in a state $s'$ preceding $s$, and no plan achieving one of the goals in $\mathcal{G} \setminus \{A\}$ deletes $A$. Goal clobbering holds if any solution plan for $A$ deletes $B$ and thus, $A < B$. Precondition violation holds if any solution for $B$ results in a deadlock from which $A$ cannot be reached anymore, i.e., again $A < B$. A composite criterion is defined that tests all three criteria simultaneously.[12] A goal $A$ is *necessarily ordered* before $B$ if it satisfies the composite criterion.

We remark that *precondition violation* seems to be equivalent to the *forced* orderings we introduced, while *goal clobbering* appears to be similar to our *reasonable* orderings. It is not possible for us to verify this conjecture as the authors of (Hüllem et al., 1999) do not give exact formal definitions. We have nothing similar to *goal subsumption* and we argue that this criterion will be rarely satisfied in natural problems: if a goal $A$ is achieved by every solution for a goal $B$ anyway, then the goal $A$ can be removed from the goal set without changing the planning task.

---

11. A plan is minimal if it contains no subplan that is also a solution plan. We remark that minimality does not mean that only shortest plans having the least number of actions are considered. In fact, minimal plans can be highly non-optimal as long as no action is truly superfluous.
12. Here, the authors are not very precise about what they mean with this. We argue that this means that two goals are ordered if they satisfy at least one of the criteria.

The authors report that they are able to detect *necessary* orderings in the artificial domains $D_iS_i$, cf. (Barrett & Weld, 1994), but fail in typical benchmark domains such as the *blocks world* or the *tyreworld*. The reason for this seems to be that their operator graphs do not represent all possible instantiations of operator schemes. As the authors claim, this makes operator graph analysis very efficient. However, the heuristic ordering $\leq_h$ that we introduced in this paper also takes almost no computation time, and succeeds in finding the goal orderings in these domains.

## 7. Outlook

Three promising avenues for future research are the following:

First, one can imagine that goal ordering information is also used *during* the search process, i.e., by not only ordering the original goal set, but also other goals that emerge during search. The major challenge seems to balance the effort on computing the goal ordering information with the savings that can result for the search process. One can easily imagine that ordering all goal sets that are ever generated can become a quite costly investment without yielding a major benefit for the planner.

Secondly, the refinement of the goal agenda with additional subgoals is another interesting future line of work. A first investigation using so-called *intermediate goals* (these are facts that the planner must make TRUE before it can achieve an original goal) has been explored inside GAM and the results are reported in (Koehler & Hoffmann, 1998). Earlier work addressing the task of learning intermediate goals can be found in (Ruby & Kibler, 1989), but this problem has not been in the focus of AI planning research since then.

A third line of work addresses the interaction of GAM with a forward-searching planning system. We have seen that GAM preserves the correctness of a planner, and that it preserves the completeness at least on deadlock-free planning domains. We have also seen, however, that solution plans using GAM can get longer, i.e., GAM does *not* preserve the optimality of a planner. Recently, planning systems that do not deliver plans of guaranteed optimality have demonstrated an impressive performance in terms of runtime and plan length, e.g., HSP, which is first mentioned in (Bonet, Loerincs, & Geffner, 1997), GRT (Refanidis & Vlahavas, 1999), and in particular FF (Hoffmann, 2000). These systems are heuristic-search planners searching forward in the state space with non-admissible, but informative heuristics.

The FF planning system developed by one of the authors has been awarded "Group A Distinguished Performance Planning System" and has also won the Schindler Award for the best performing planning system in the Miconic 10 Elevator domain (ADL track) at the AIPS 2000 planning competition. The integration of goal agenda techniques into the planner is one of the factors that enabled the excellent behavior of FF in the competition: they were crucial for scaling to *blocks world* problems of 50 blocks, helped by about a factor 2 on *schedule* and *Miconic 10*, and never slowed down the algorithm.

Forward state-space search is a quite natural framework to be driven by the goal agenda: Simply let the planner solve a subproblem, and start the next search from the state where the last search ended. Even more appealing, heuristic forward-search planners have a deeper kind of interaction with GAM than for example GRAPHPLAN-style planners. In addition to the smaller problems they are facing when using the goal agenda, their heuristics are influenced because they employ techniques for estimating the goal distance from a state. When using the goal agenda, different goal sets result at each stage of the planning process and therefore, the goal-distance estimate will be different, too. Currently a heuristic device inside the FF search algorithm is being developed, which knows that it is being driven by a goal agenda, and which has access to the complete set of goals. This information can be used to further prune unpromising branches from the search space when it discovers that currently achieved goals will probably have to be destroyed and reachieved later on.